\documentclass[preprint]{elsarticle}
\usepackage{amsmath,amsfonts}
\usepackage{amsthm}
\usepackage{algorithmic}
\usepackage{algorithm}
\usepackage{array}
\usepackage{textcomp}
\usepackage{stfloats}
\usepackage{url}
\usepackage{verbatim}
\usepackage{graphicx}
\usepackage{multirow}
\usepackage{bbding}
\usepackage{gensymb}
\usepackage{booktabs}
\usepackage{subfigure}
\usepackage{color}
\theoremstyle{plain}
\newtheorem{theorem}{Theorem}
\theoremstyle{remark}

\usepackage{lineno,hyperref}

\begin{document}

\begin{frontmatter}

\title{Seeking Commonness and Inconsistencies: A Jointly Smoothed Approach to Multi-view Subspace Clustering}

\author[1]{Xiaosha Cai}
\ead{xiaoshacai@hotmail.com}
\author[1]{Dong Huang\corref{cor1}}
\ead{huangdonghere@gmail.com}
\author[1]{Guang-Yu Zhang}
\ead{guangyuzhg@foxmail.com}
\author[2,3]{Chang-Dong Wang}
\ead{changdongwang@hotmail.com}
\address[1]{College of Mathematics and Informatics, South China Agricultural University, China}
\address[2]{School of Computer Science and Engineering, Sun Yat-sen University, China}
\address[3]{Guangdong Key Laboratory of Information Security Technology, China}
\cortext[cor1]{Corresponding author}

\begin{abstract}
 Multi-view subspace clustering aims to discover the hidden subspace structures from multiple views for robust clustering,  and has been attracting considerable attention in recent years. Despite significant progress, most of the previous multi-view subspace clustering algorithms are still faced with two limitations. First, they usually focus on the consistency (or commonness) of multiple views, yet often lack the ability to capture the cross-view inconsistencies in subspace representations. Second, many of them overlook the local structures of multiple views and cannot jointly leverage multiple local structures to enhance the subspace representation learning. To address these two limitations, in this paper, we propose a \textbf{j}ointly \textbf{s}moothed \textbf{m}ulti-view subspace \textbf{c}lustering (JSMC) approach. Specifically, we simultaneously incorporate the cross-view commonness and inconsistencies into the subspace representation learning. The view-consensus grouping effect is presented to jointly exploit  the local structures of multiple views to regularize the view-commonness representation, which is further associated with the low-rank constraint via the nuclear norm to strengthen its cluster structure. Thus the cross-view commonness and inconsistencies, the view-consensus grouping effect, and the low-rank representation are seamlessly incorporated into a unified objective function, upon which an alternating optimization algorithm is performed to achieve a robust  subspace representation for clustering. Experimental results on a variety of real-world multi-view datasets confirm the superiority of our approach. Code available: \url{https://github.com/huangdonghere/JSMC}.

\end{abstract}


\end{frontmatter}


\section{Introduction}
Multi-view data widely exist in many real-world scenarios, where the data instances may have multiple feature sets collected from different sources or views. For example, a piece of news can be reported in different languages, such as English, French, and Spanish. A webpage can be described in terms of videos, texts, and images. Different data views can provide rich and versatile information for exploring the hidden structure of the data. But in the meantime the multi-view data have also brought new challenges to the field of data mining and knowledge discovery. Various studies have been conduted in multi-view data analytics, where the multi-view clustering problem has been attracting increasing attention in recent years \cite{chao21_tai}.

When considering multi-view clustering, a na\"ive strategy is to concatenate the features of multiple views and then perform some traditional single-view clustering algorithms on the concatenated features, which, however, ignores the consensus and complementary information among multiple views and is rarely adopted in practice. To exploit the rich information of multiple views, many multi-view clustering algorithms have been developed in the literature \cite{chao21_tai},
which can be divided into the co-training based methods \cite{Kumar2011cotrain,kumar2011co,Ye2016}, the multi-kernel based methods  \cite{Tzor12_icdm,guo14_icpr,zhang21_INS,zhou2020subspace}, the graph learning based methods \cite{nie17_ijcai,ZhanFusion,Zhan2018,liangTNNLS}, the deep learning based methods \cite{xie2021joint,wang22_tbd}, and the subspace learning based methods \cite{gao2015multi,wang2016iterative,zhang2018generalized}. 
Among the existing multi-view clustering methods, the multi-view subspace clustering has been an important category~\cite{gao2015multi,wang2016iterative,wang2020smoothness, zhang2018generalized,lv2021multi,zhang2021joint,zhang2020one}. Multi-view subspace clustering aims to uncover the low-dimensional subspaces from multiple views, and represent the data instances in these subspaces for clustering analysis.
For example, Gao et al.~\cite{gao2015multi} proposed to learn multiple subspace representations on multiple views and further constrain them to be consistent via a common cluster indicator.
Wang et al.~\cite{wang2016iterative} exploited the graph Laplacian regularized low-rank representation in multi-view subspace learning which incorporates the manifold information of each view. Lv et al. \cite{lv2021multi} considered the multi-view subspace clustering problem in the partition space and jointly learned a graph for each view, a partition for each view, and a final consensus partition.

The existing multi-view subspace clustering algorithms \cite{gao2015multi,wang2016iterative,wang2020smoothness, zhang2018generalized,lv2021multi,zhang2021joint,zhang2020one} typically learn multiple subspace representations from multiple views and then constrain the multiple representations by incorporating some consistency regularization. Despite the significant progress that has been achieved, there are still two common limitations to the prior works. First, they mostly focus on the consistency (or commonness), yet often lack the ability to explicitly capture the cross-view inconsistency in the subspace representations. Second, they often neglect the local structures of multiple views, and cannot jointly exploit the multiple local structures to assist the subspace learning and the final clustering.

More recently, some efforts have been made to partially address the above two limitations. Tang et al. \cite{tang19_tmm} learned a joint affinity matrix for multi-view subspace clustering with the diversity regularization and a rank constraint. Luo et al. \cite{luo2018consistent} incorporated consistency and specificity into multi-view subspace clustering. However, these works \cite{tang19_tmm,luo2018consistent} still ignore the rich information hidden in multi-view local structures, which may significantly benefit the robustness of the subspace representation learning.
In the single-view scenario, Hu et al.~\cite{hu2014smooth} incorporated the smooth regularization into the subspace clustering framework, which is able to leverage the local (or neighborhood) structure in the subspace learning via the grouping effect. Chen et al.~\cite{chen2020multiview} extended the single-view grouping effect into the multi-view scenario, and proposed the multi-view subspace clustering with grouping effect algorithm. Despite this, on the one hand, the algorithm in \cite{chen2020multiview} only considered the grouping effect of each view separately, but cannot jointly enforce the multi-view grouping effect (or view-consensus grouping effect) on a unified representation. On the other hand, it still fails to jointly consider the cross-view consistency and inconsistency in the multi-view subspace learning.

\begin{figure*}[!t]
	\begin{center}
		{
			{\includegraphics[width=1\columnwidth]{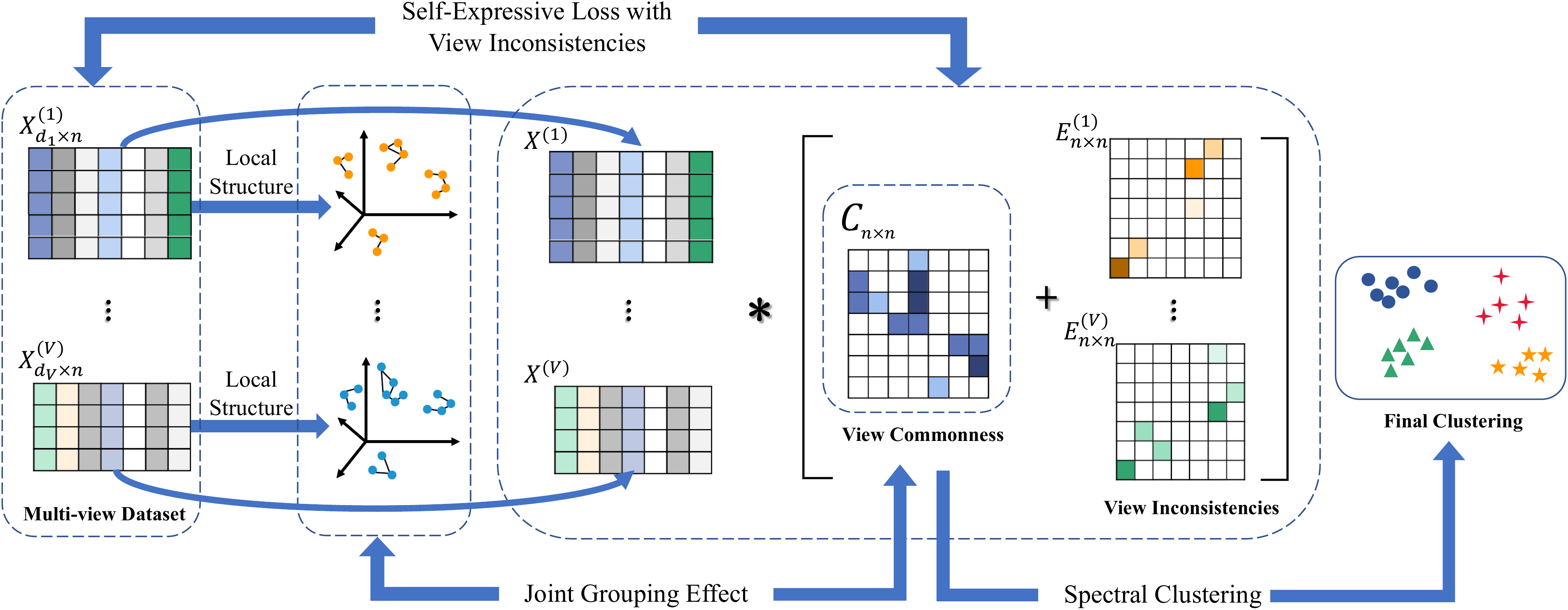}}}
		\caption{Flow diagram of the proposed JSMC approach. Given a set of data instances with $V$ views, namely, $X^{(1)}, X^{(2)}, \ldots, X^{(V)} $, our approach decomposes the subspace representation on the view $X^{(v)}$ into a view-commonness matrix $C$ and a view-inconsistency matrix $E^{(v)}$, based on which the self-expressive loss with view inconsistencies is incorporated to learn the subspace representation and the joint grouping effect is enforced on the view-commonness matrix. With the common representation learned, the spectral clustering can be utilized to obtain the final clustering result.}
		\label{fig:flowchart}
	\end{center}
\end{figure*}

To simultaneously address the above two limitations, in this paper,  we propose a novel \textbf{j}ointly \textbf{s}moothed \textbf{m}ulti-view subspace \textbf{c}lustering (JSMC) approach. In comparison with previous works, our JSMC approach is able to integrates (i) the \emph{cross-view commonness and inconsistencies} in subspace representations,  (ii) the jointly smoothed regularization via \emph{view-consensus grouping effect}, and (iii) the \emph{low-rank representation} into a unified multi-view subspace framework. In particular, to explicitly model the cross-view commonness and inconsistencies, we decompose the subspace representation on each view into a common part and an inconsistent part. In contrast with the conventional practice of extracting consistent parts from multiple views and then constraining these consistent parts to be similar \cite{gao2015multi} , the commonness is a more strict concept than the consistency. Especially, we aim to discover a common subspace representation across multiple views while simultaneously capturing the view-particularities via the inconsistent parts. To leverage the local structures of multiple views, the view-consensus grouping effect is formulated between the multi-view $K$-nearest neighbor ($K$-NN) graphs and a common subspace representation, which enforces the multi-view smooth regularization on the subspace representation learning. Furthermore, the low-rank constraint is incorporated to enhance the cluster structure in the common subspace representation.
With the cross-view consistency and inconsistencies, the view-consensus grouping effect, and the low-rank constraint jointly modeled into a unified objective function, an alternating minimization algorithm is then designed to perform the optimization and learn a robust subspace representation, upon which the spectral clustering can be utilized to achieve the final clustering result. Experiments on eight real-world multi-view datasets demonstrate the superiority of the proposed JSMC approach.

For clarity, the main contributions of this work are summarized as follows:
\begin{itemize}
\item We simultaneously and explicitly model the cross-view commonness and inconsistencies to assist the multi-view subspace clustering, which adaptively learns the common subspace representation while capturing view particularities.
\item We extend the view-specific grouping effect to view-consensus grouping effect, which together with the inconsistency-aware subspace learning and the low-rank representation learning constitute our unified objective function for robust multi-view subspace clustering.
\item A novel multi-view subspace clustering approach termed JSMC is proposed. Extensive experimental results on a variety of multi-view datasets have demonstrated the superior performance of the proposed approach.
\end{itemize}

The rest of the paper is organized as follows. We review the related work on multi-view clustering in Section~\ref{sec:related work}. The proposed JSMC approach is described in Section~\ref{sec:framework}. Experimental results are reported in Section~\ref{sec:experimet}. Finally, the conclusion of the paper is provided in Section~\ref{sec:conclusion}.

\section{Related work}
\label{sec:related work}
Multi-view clustering aims to utilize the features of multiple views to achieve a unified clustering result.
In recent years, many multi-view clustering methods have been designed from different technical perspectives \cite{Kumar2011cotrain,kumar2011co,Ye2016,Tzor12_icdm,guo14_icpr,zhang21_INS,zhou2020subspace,nie17_ijcai,ZhanFusion,Zhan2018,gao2015multi,wang2016iterative,zhang2018generalized,steffen2004multiview,zhao2017deeep,chaudhuri2009cca,yao19_ijcai,yan2020_if,huang21_nn,liang19_icdm,hu20_tcyb,hu22_tip}. Some important categories in multi-view clustering include the co-training based methods \cite{Kumar2011cotrain,kumar2011co,Ye2016}, the multi-kernel based methods  \cite{Tzor12_icdm,guo14_icpr,zhang21_INS,zhou2020subspace}, the graph learning based methods \cite{nie17_ijcai,ZhanFusion,Zhan2018,liangTNNLS,liang19_icdm}, the deep learning based methods \cite{xie2021joint,wang22_tbd}, and the subspace learning based methods \cite{gao2015multi,wang2016iterative,zhang2018generalized}.

The co-training based methods \cite{Kumar2011cotrain,kumar2011co,Ye2016,ling2018cotraining} aim to learn a clustering result by minimizing the multi-view disagreement.
Kumar et al.~\cite{Kumar2011cotrain} proposed a co-training method where each cluster was able to utilize the complementary information learned from other clusters, and applied the co-training method to multi-view spectral clustering.
Further, Kumar et al.~\cite{kumar2011co} proposed a multi-view spectral clustering method with co-regularization, which imposed the regularization on the eigenvectors of the graph Laplacians constructed from multiple views.
Ye et al.~\cite{Ye2016} designed a co-regularized kernel $k$-means method that automatically learned the weights of different views for robust clustering.

The multi-kernel based methods \cite{Tzor12_icdm,guo14_icpr,zhang21_INS,zhou2020subspace,sally2022kernel} express multiple views in terms of  multiple kernel matrices and combine the multiple kernels to achieve the final clustering. Tzortzis and Likas \cite{Tzor12_icdm} proposed a weighted combination scheme for multiple kernels, which iteratively updated the kernel weights and  recomputed the cluster labels to optimize the clustering result. Guo et al. \cite{guo14_icpr} proposed a multiple kernel learning method for multi-view spectral clustering, where the kernel matrix learning and the spectral clustering optimization were combined in a unified framework. Zhang et al. \cite{zhang21_INS} designed a multi-view clustering method based on multiple kernel low-rank representation, which combined the co-regularization with the low-rank multiple kernel trick.

The graph learning based methods \cite{nie17_ijcai,ZhanFusion,Zhan2018,liangTNNLS,liang19_icdm} seek to learn a unified graph by fusing multiple affinity graphs from multiple views. Upon the unified graph,  the final clustering can be obtained via some graph partitioning algorithm. Nie et al. \cite{nie17_ijcai} took the importance of views into account and developed a self-weighted scheme to fuse the multiple graphs. Zhan et al. \cite{Zhan2018} learned a consensus graph by minimizing the disagreement between the graphs from different views and incorporating the rank constraint on the Laplacian matrix. Liang et al. \cite{liangTNNLS} considered the consistency and inconsistency in a graph learning framework, and proposed two graph fusion algorithms, termed similarity graph fusion (SGF) and dissimilarity (distance) graph fusion (DGF), respectively.

The deep learning based methods \cite{xie2021joint,wang22_tbd} learn the feature representations of multiple views via deep neural networks and typically enforce some consistency constraint on the multi-view representations for obtaining the final clustering. For example, Xie et al. \cite{xie2021joint} proposed the deep multi-view joint clustering (DMJC) method to learn multi-view feature representations and in the meantime utilize the complementary information among multiple views. Wang et al. \cite{wang22_tbd} designed a generalized deep learning multi-view clustering (GDLMC) method based on nonnegative matrix factorization (NMF).

The subspace learning based methods leverage the self-expressive loss (or reconstruction loss) to discover the low-dimensional subspaces from multiple views for improving representation learning and clustering. Gao et al.~\cite{gao2015multi} presented a multi-view subspace clustering method with the multiple subspace representations constrained to be consistent.
Wang et al.~\cite{wang2016iterative} incorporated the manifold information of each view by adopting the graph Laplacian regularized low-rank representation in multi-view subspace learning. Zhang et al. \cite{zhang2018generalized} learned a shared multi-view latent representation, which is able to depict the data instances more comprehensively than individual views and make the subspace representation more robust.

\section{Jointly Smoothed Multi-view Subspace Clustering}
\label{sec:framework}
In this section, we describe the proposed JSMC approach in detail. Specifically, Section~\ref{sec:inconsistecy_in_subspace_learning} presents the incorporation of the view-commonness and view-inconsistencies into the multi-view subspace learning model. Section~\ref{sec:joint_grouping_effect} extends the view-specific grouping effect to the view-consensus grouping effect for jointly smoothed regularization. Section~\ref{sec:overall_objective} describes the  formulation of the unified objective function, which will be optimized via an alternating minimization algorithm in Section~\ref{sec:opt}. Finally, the computational complexity of the proposed JSMC approach is analyzed in Section~\ref{sec:complexity}.

\subsection{Multi-view Inconsistency in Subspace Learning}
\label{sec:inconsistecy_in_subspace_learning}
In this section, we present the formulation of the multi-view subspace learning with cross-view commonness and inconsistencies. In subspace learning, it is generally assumed that the structure of high-dimension data can be represented by a union of low-dimension subspaces.
According to the self-expressiveness property~\cite{elhamifar2013sparse}, a data instance can be represented as a linear combination of other instances. Given a (single-view) dataset $X=\left[\mathbf{x}_{1}, \mathbf{x}_{2}, \ldots, \mathbf{x}_{n}\right] \in \mathbb{R}^{d \times n}$, where $\mathbf{x}_{i}\in \mathbb{R}^{d}$ is the $i$-th instance,  $n$ is the number of instances in the dataset, and $d$ is the dimension, the traditional subspace learning based clustering (or subspace clustering for short) methods typically seek to learn a self-representation matrix by the self-expressiveness property~\cite{elhamifar2013sparse}, such that
\begin{align}
	\label{eq:Self-represent}
	X=X Z+A,
\end{align}
where $A \in \mathbb{R}^{n \times n}$ is the  error matrix, $Z=\left[\mathbf{z}_{1}, \mathbf{z}_{2}, \ldots, \mathbf{z}_{n}\right]\in \mathbb{R}^{n \times n}$ is the self-representation matrix, and $\mathbf{z}_{i}$ is a representation vector of data instance $\mathbf{x}_{i}$. The subspace clustering aims to find a better representation by minimizing the self-express error $A$, whose objective function can be formulated as
\begin{align}
\label{eq:SubspaceClustering}
\underset{Z}{\text{min}} & \|X-X Z\|_{F}^{2}+\alpha \Omega(Z), \\
\text { s.t. } &\quad z_{ij}=0, Z^{\top} \mathbf{1}=\mathbf{1},\nonumber
\end{align}
where $z_{ij}$ denotes the $(i,j)$-th entry of $Z$, $\Omega(Z)$ denotes a regularization term on $Z$, and $\alpha\geq 0$ is a hyper-parameter to control the influence of the regularization term. By constraining $z_{ij}=0$, it enforces that each instance $\mathbf{x}_{i}$ can only be represented by other instances. By optimizing the objective~(\ref{eq:SubspaceClustering}), the self-representation (i.e., subspace representation) $Z$ can be learned, upon which the symmetrical affinity matrix can be constructed as $W=(|Z|+\left|Z^{\top}\right|)/2$ for the final spectral clustering process~\cite{hu2014smooth}, where $|\cdot|$ is the absolute operator.

In the multi-view scenarios, the (single-view) subspace clustering formulation can be extended to multiple views by learning a self-representation matrix for each view and then enforcing some consistency constraint on the multiple representations. For most of multi-view subspace clustering algorithms~\cite{gao2015multi,wang2016iterative,yin2015multi,kang2020partition}, they obtain a unified representation matrix or explore a latent space for all views according to the consensus principle. Given a multi-view dataset $X=\{X^{(1)}, X^{(2)}, \ldots, X^{(V)} \} \in \mathbb{R}^{d \times n}$, where $X^{(v)} \in \mathbb{R}^{d_v \times n}$ is the $v$-th view, $d_v$ is the dimension of the $v$-th view. It holds that $d=d_1 + \cdots + d_V$. Then the objective function of the multi-view subspace clustering can be formulated as
\begin{align}
&\min_{Z^{(1)},\cdots,Z^{(V)}} \sum_{v=1}^V\left\|X^{(v)}-X^{(v)} Z^{(v)}\right\|_{F}^{2}+\alpha \Omega(Z^{(1)},\cdots,Z^{(V)}), \\
&\text { s.t. } \quad z^{(v)}_{ij}=0, (Z^{(v)})^{\top} \mathbf{1}=\mathbf{1}, \text{~for~} v = 1, \cdots,V,\nonumber
\end{align}
where $z^{(v)}_{ij}$ denotes the $(i,j)$-th entry of the self-representation matrix $Z^{(v)}$ of the $v$-th view, and $\Omega(Z^{(1)},\cdots,Z^{(V)})$ denotes the regularization term which typically constrains the consistency of the multiple views. However, the prior works~\cite{gao2015multi,wang2016iterative,yin2015multi,kang2020partition} mostly focus on the consistency of the multi-view subspace representations, yet often lack the ability to explicitly capture the inconsistency between different views, which may undermine their ability to make full use of versatile information of multiple views. In light of this, we can decompose the subspace representation $Z^{(v)}$ into a view-consistency matrix $C^{(v)}$ and a view-inconsistency matrix $E^{(v)}$, that is
\begin{align}
	\label{eq:decompose_consistent}
	Z^{(v)}=C^{(v)}+E^{(v)}.
\end{align}

Then, the objective function of multi-view subspace clustering can be re-formulated as
\begin{align}
	\label{eq:objective_consistent_inconsistent}
		\underset{C^{(1)},\cdots,C^{(V)},E^{(1)},\cdots,E^{(V)}}{\text{min}} &\sum_{v=1}^{V}\|X^{(v)}-X^{(v)}(C^{(v)}+E^{(v)})\|_F^2 \\
		& + \alpha \Omega_1(C^{(1)},\cdots,C^{(V)}) + \beta\Omega_2(E^{(1)},\cdots,E^{(V)})\nonumber
\end{align}
where $\Omega_1(C^{(1)},\cdots,C^{(V)})$ and $\Omega_2(E^{(1)},\cdots,E^{(V)})$ are regularization terms on the view-consistency matrices and the view-inconsistency matrices, respectively, and $\alpha\geq 0$ and $\beta\geq 0$ are hyper-parameters.

Further, we aim to extend the concept of \emph{view-consistency} to a more strict concept of \emph{view-commonness}. Specifically, the consistency typically requires multiple representations (from multiple views) to be similar, e.g., by enforcing the similarity between the $V$ representations ranging from $C^{(1)}$ to $C^{(V)}$ via the regularization term $\Omega_1(\cdot)$ in objective~\eqref{eq:objective_consistent_inconsistent},
while the concept of commonness goes one step further by extracting a \emph{common} part from multiple representations. Formally, we can decompose each subspace representation matrix $Z^{(v)}$ into a view-commonness matrix $C$ and a view-inconsistency matrix $E^{(v)}$, that is
\begin{align}
	\label{eq:decompose_common}
	Z^{(v)}=C+E^{(v)}.
\end{align}
where the view-commonness matrix $C$ is shared by all views and the view-inconsistency matrices $E^{(1)},\cdots,E^{(V)}$ explicitly capture the inconsistencies in multiple views. Note that the inconsistency is a broader concept than noise, which may come from noise or view-specific characteristics.
Thereafter, the objective function can be rewritten as
\begin{align}
	\label{eq:objective_common_inconsistent}
	\underset{C,E^{(1)},\cdots,E^{(V)}}{\text{min}} &\sum_{v=1}^{V}\|X^{(v)}-X^{(v)}(C+E^{(v)})\|_F^2 \\
	& + \alpha \Omega_1(C) + \beta\Omega_2(E^{(1)},\cdots,E^{(V)})\nonumber
\end{align}
Notably, the objective~(\ref{eq:objective_common_inconsistent}) can be viewed as a special instance of the objective~(\ref{eq:objective_consistent_inconsistent}) by additionally constraining that $C^{(1)}=C^{(2)}=\cdots =C^{(V)}$.

\subsection{From View-Specific to View-Consensus Grouping Effect}
\label{sec:joint_grouping_effect}

Previous multi-view subspace clustering works \cite{gao2015multi,wang2016iterative,zhang2018generalized}  often lack the ability to explicitly preserve the locality in the learned subspace representations. To explain this issue, we first discuss it in the single-view scenario. In terms of locality preservation, if two instances $\mathbf{x}_{i}$ and $\mathbf{x}_{j}$ are similar,  their subspace representations $\mathbf{z}_{i}$ and $\mathbf{z}_{j}$ should also be close to each other, which is called the grouping effect, also known as the smooth regularization \cite{hu2014smooth}. Formally, the grouping effect is defined in Definition 1.

\textbf{Definition 1\quad (Grouping Effect).} \emph{Given a set of data instances $X=\left[\mathbf{x}_{1}, \mathbf{x}_{2}, \ldots, \mathbf{x}_{n}\right] \in \mathbb{R}^{d \times n}$, a self-representation matrix $Z=\left[\mathbf{z}_{1}, \mathbf{z}_{2}, \ldots, \mathbf{z}_{n}\right] \in \mathbb{R}^{n \times n}$ has the grouping effect if $\left\|\mathbf{x}_{i}-\mathbf{x}_{j}\right\|_{2} \rightarrow 0 \Rightarrow\left\|\mathbf{z}_{i}-\mathbf{z}_{j}\right\|_{2} \rightarrow 0, \forall i \neq j$.}

The smooth regularization (via grouping effect) has shown its advantage in locality preserving for single-view clustering \cite{hu2014smooth}. Recently Chen et al. \cite{chen2020multiview} incorporated the grouping effect into the multi-view subspace clustering in a view-specific manner, which enforces the grouping effect on each view separately but still cannot exploit the grouping effect of multiple views in a jointly smoothed formulation.

From the view-specific perspective, for the $v$-th view, if the subspace representation $Z^{(v)}$ is decomposed into a view-consistency representation $C^{(v)}$ and a view-inconsistency representation $E^{(v)}$ (as shown in Eq.~(\ref{eq:decompose_consistent})), then we can enforce the view-specific grouping effect by considering the affinity (or similarity) of two instances $\mathbf{x}^{(v)}_i$ and $\mathbf{x}^{(v)}_j$ and their view-consistency representations $\mathbf{c}^{(v)}_i$ and $\mathbf{c}^{(v)}_j$. More specifically, for the $v$-th view, if $\mathbf{x}^{(v)}_i$ and  $\mathbf{x}^{(v)}_j$ are similar, then $\mathbf{c}^{(v)}_i$ and $\mathbf{c}^{(v)}_j$ should also be close to each other, which can be achieved by optimizing the following objective:
\begin{align}
	\label{eq:view-specific-group-effect}
\underset{C^{(v)}}{\text{min}}   \frac{1}{2} \sum_{i=1}^{n} \sum_{j=1}^{n} w_{i j}^{(v)}\left\|\mathbf{c}_{i}^{(v)}-\mathbf{c}_{j}^{(v)}\right\|_{2}^{2},
\end{align}
where $W^{(v)}$ denotes the affinity matrix (or similarity matrix) with $w_{i j}^{(v)}$ being its $(i,j)$-th entry. In general, the affinity matrix $W^{(v)}$ can be defined by constructing a fully-connected graph, an $\varepsilon$-neighbor graph, or a $K$-nearest neighbor ($K$-NN) graph. In this work, the $K$-NN graph is adopted to preserve the local structure of the data. Let $L^{(v)}=D^{(v)}-W^{(v)}$ be the Laplacian matrix of $W^{(v)}$, where $D^{(v)}$ is the degree matrix with its $(i,j)$-th entry being $d_{ii}^{(v)}=\sum_{j} w_{i j}^{(v)}$. Then the objective (\ref{eq:view-specific-group-effect}) can be rewritten as
\begin{align}
	\label{eq:view-specific-group-effect-trace}
	\underset{C^{(v)}}{\text{min}}   \operatorname{Tr}\left(C^{(v)} L^{(v)} (C^{(v)})^{\top}\right).
\end{align}

To extend from view-consistency to view-commonness (as discussed in Section~\ref{sec:inconsistecy_in_subspace_learning}), we decompose the subspace representation $Z^{(v)}$ into a view-commonness matrix $C$ and a view-inconsistency matrix $E^{(v)}$. To jointly enforce the grouping effect, we expect that the local structures in different views can jointly influence the learning of the common representation matrix $C$, which gives rise to the following objective:
\begin{align}
	\label{eq:view-common-group-effect}
		\underset{C}{\text{min}}    \frac{1}{2}\sum_{v=1}^{V}\sum_{i=1}^{n}\sum_{j=1}^{n}w^{(v)}_{ij}\|\mathbf{c}_{i}-\mathbf{c}_{j}\|_2^2,
\end{align}
where $\mathbf{c}_{i}$ and $\mathbf{c}_{j}$ denote the common representations of the $i$-th and $j$-th instances, respectively.  Further, the objective (\ref{eq:view-common-group-effect}) can be rewritten as
\begin{align}
	\label{eq:view-common-group-effect-trace}
	\underset{C}{\text{min}}     \sum_{v=1}^{V}\operatorname{Tr}(CL^{(v)}C^{\top}).
\end{align}

Different from the conventional view-specific grouping effect \cite{chen2020multiview}, through the objective~\ref{eq:view-common-group-effect}, we enforce the grouping effect in a view-consensus manner, which collectively explores the local structures of multiple views to assist the learning of the multi-view common representation.

\subsection{Overall Objective Function}
\label{sec:overall_objective}

With the cross-view consistency and inconsistency captured in objective~(\ref{eq:objective_common_inconsistent}) and the view-consensus grouping effect exploited in objective~(\ref{eq:view-common-group-effect-trace}),
in this section, we further specify the regularization terms on the view-commonness and view-inconsistency matrices, and present the overall objective function of the proposed approach.

For the view-commonness matrix, i.e., the common representation matrix $C$, we take advantage of the low rank constraint via the nuclear norm, which reduces the redundancy of the matrix and enhances the cluster structure \cite{zhang2015low}. That is
\begin{align}
	\label{eq:nuclear_norm}
	\underset{C}{\text{min}} \| C \|_* .
\end{align}

For the view-inconsistency matrices $E^{(1)},\cdots,E^{(V)}$, the Frobenius norm is utilized to regularize them. That is

\begin{align}
	\label{eq:inconsistency_F_norm}
	\underset{E^{(1)}, \ldots, E^{(V)}}{\text{min}}  \sum_{v=1}^{V}{\|E^{(v)}\|_F^2}.
\end{align}

From objectives (\ref{eq:objective_common_inconsistent}), (\ref{eq:view-common-group-effect-trace}), (\ref{eq:nuclear_norm}), and (\ref{eq:inconsistency_F_norm}), the overall objective function can be formulated as follows:

\begin{equation}
	\begin{aligned}
		\label{eq:ObjFunction}
		\underset{C,E^{(1)}, \ldots, E^{(V)}}{\text{min}}&\mathcal{O}\left(C,E^{(1)}, \ldots, E^{(V)}\right)= \\ &\underbrace{\sum_{v=1}^{V}\|X^{(v)}-X^{(v)}(C+E^{(v)})\|_F^2}_{\text{Inconsistency-Aware Self-Expressive Loss}} + \underbrace{\alpha \sum_{v=1}^{V} \operatorname{Tr}(CL^{(v)}C^{\top})}_{\text{Jointly Smoothed Representation}}\\
		&+\underbrace{\beta \sum_{v=1}^{V}{\|E^{(v)}\|_F^2}}_{\text{View-Inconsistency Regularization}} + \underbrace{\lambda \| C \|_* }_{\text{Low-Rank Regularization}}
	\end{aligned}
\end{equation}

With this objective function, the cross-view consistency and inconsistency, the view-consensus grouping effect, and the low rank representation are simultaneously leveraged. With the common representation $C$ learned, the spectral clustering can then be utilized to obtain the final clustering. In the following section, we will present the optimization of the unified objective function.

\subsection{Optimization}
\label{sec:opt}
In this section, we optimize the objective function~(\ref{eq:ObjFunction}) by using an alternating minimizing algorithm. Before solving this problem, to make the problem separable, we introduce an auxiliary variable $S$ for the common representation matrix $C$ such that $S=C$. Thus, the objective (\ref{eq:ObjFunction}) can be rewritten as follows:
\begin{equation}
\begin{aligned}
\underset{S,C,E^{(1)}, \ldots, E^{(V)}}{\text{min}} &\mathcal{O}\left(S,C,E^{(1)}, \ldots, E^{(V)}\right)=\\
&\sum_{v=1}^{V}\|X^{(v)}-X^{(v)}(S+E^{(v)})\|_F^2 + \alpha \sum_{v=1}^{V} \operatorname{Tr}(SL^{(v)}S^{\top})\\ &+\beta \sum_{v=1}^{V}{\|E^{(v)}\|_F^2} + \lambda \| C \|_* \\
&\text { s.t. } \quad S=C
\end{aligned}
\end{equation}
Then we have the corresponding augmented Lagrangian function:
\begin{equation}
	\label{eq:object_Lagr}
\begin{aligned}
\underset{S,C,E^{(1)}, \ldots, E^{(V)}}{\text{min}} &\mathcal{L}\left(S,C,E^{(1)}, \ldots, E^{(V)}\right)= \\ & \sum_{v=1}^{V}\|X^{(v)}-X^{(v)}(S+E^{(v)})\|_F^2 + \alpha \sum_{v=1}^{V} \operatorname{Tr}(SL^{(v)}S^{\top})\\
&+\beta \sum_{v=1}^{V}{\|E^{(v)}\|_F^2} + \lambda \| C \|_* + \frac{\mu}{2} \|S-C+\frac{Y}{\mu}\|_F^2\\
\end{aligned}
\end{equation}
where $\mu>0$ is the penalty parameter, and $Y$ is the Lagrangian multiplier. Next, each variable can be iterative updated with other variables fixed \cite{fang2020dynamic}.

\subsubsection{\textbf{$S$-Subproblem}}

With $E^{(1)}, \ldots, E^{(V)}$ and $C$ fixed, the variable $S$ is updated. Specifically, the objective (\ref{eq:object_Lagr}) can be rewritten as follows:
\begin{equation}
\begin{aligned}
\underset{S}{\text{min}} \mathcal{L}\left(S\right)&= \sum_{v=1}^{V}\|X^{(v)}-X^{(v)}(S+E^{(v)})\|_F^2 \\
& + \alpha \sum_{v=1}^{V} \operatorname{Tr}(SL^{(v)}S^{\top})+ \frac{\mu}{2} \|S-C+\frac{Y}{\mu}\|_F^2\\
\end{aligned}
\end{equation}
Then we obtain its partial derivative w.r.t. $S$, that is
\begin{equation}
\begin{aligned}
\frac{\partial \mathcal{L}(\mathrm{S})}{\partial S}&=\sum_{v=1}^{V}2X^{(v)^{\top}} X^{(v)}\left(-I+ S+ E^{(v)}\right)  \\
&+2\alpha \sum_{v=1}^{V} S L^{(v)}+\frac{\mu}{2}\left(2 S-2 C+\frac{2}{\mu} Y\right),\\
\end{aligned}
\end{equation}
where $I$ is an identity matrix. By setting $\frac{\partial \mathcal{L}(\mathrm{S})}{\partial S}=0$, we have
\begin{equation}
\begin{aligned}
\label{eq:Sproblem}
\left(2 \sum_{v=1}^{V} X^{(v)^{\top}} X^{(v)}+\mu I\right) S+S * 2 \alpha \sum_{v=1}^{V} L^{(v)} \\
=\sum_{v=1}^{V}\left[2 X^{(v)^{\top}} X^{(v)}\left(I-E^{(v)}\right)\right]+\mu C-Y
\end{aligned}
\end{equation}
The problem in Eq.~(\ref{eq:Sproblem}) is a standard Sylvester equation. Therefore, the Bartels-Stewart algorithm~\cite{bartels1972solution} can be utilized to solve this problem.

\subsubsection{\textbf{$C$-Subproblem}}

With $E^{(1)}, \ldots, E^{(V)}$ and $S$ fixed, the variable $C$ is updated. The objective (\ref{eq:object_Lagr}) w.r.t. the common representation matrix $C$ can be rewritten as follows:
\begin{equation}
    \begin{aligned}
        \label{eq:originalC}
        &\underset{C}{\text{min}} \quad\mathcal{L}\left(C\right)= \lambda \| C \|_* + \frac{\mu}{2} \|S-C+\frac{Y}{\mu}\|_F^2\\
    \end{aligned}
\end{equation}
By solving this problem via the singular value thresholding (SVT) operator~\cite{cai2010singular}, we have
\begin{equation}
    \label{eq:Cproblem}
    C=U_{C} \delta_{\frac{\lambda}{\mu}}(\Sigma_C) V_{C}^{\top},
\end{equation}
where $U_{C} \Sigma_C V_{C}^{\top}$ denotes the singular value decomposition (SVD) of $S+\frac{Y}{\mu}$, and $\delta_{\frac{\lambda}{\mu}}(\cdot)$ denotes the  shrinkage operator, which can be defined as
\begin{equation}
    \begin{aligned}
         \delta_{\frac{\lambda}{\mu}}(\Sigma_C)=\max \left(0, \Sigma_C-\frac{\lambda}{\mu}\right)+\min \left(0, \Sigma_C+\frac{\lambda}{\mu}\right)
    \end{aligned}
\end{equation}

\subsubsection{\textbf{$E^{(v)}$-Subproblem}}

With $C$ and $S$ fixed, the variables $E^{(1)}, \ldots, E^{(V)}$ are updated. The objective (\ref{eq:object_Lagr}) w.r.t. the view-inconsistency matrices can be rewritten as
\begin{equation}
    \begin{aligned}
        \label{eq:Eupdate}
        &\underset{E^{(1)}, \ldots, E^{(V)}}{\text{min}} \mathcal{L}\left(E^{(1)}, \ldots, E^{(V)}\right)= \\ &\quad \sum_{v=1}^{V}\|X^{(v)}-X^{(v)}(S+E^{(v)})\|_F^2 +\beta \sum_{v=1}^{V}{\|E^{(v)}\|_F^2}
    \end{aligned}
\end{equation}
In particular, for each view $v$, its view-inconsistency matrix $E^{(v)}$ can be updated respectively. Then the objective~(\ref{eq:Eupdate}) can be reformulated as follows:
\begin{equation}
    \begin{aligned}
        \label{eq:Eobj}
        &\underset{E^{(v)}}{\text{min}} \mathcal{L}\left(E^{(v)}\right)=\|X^{(v)}-X^{(v)}(S+E^{(v)})\|_F^2 +\beta {\|E^{(v)}\|_F^2}
    \end{aligned}
\end{equation}
By setting the derivative of the above objective function to $0$, $E^{(v)}$ can be updated as follows:
\begin{equation}
    \label{eq:Eproblem}
    E^{(v)}=\left(X^{(v)^{\top}} X^{(v)}+\beta I\right)^{-1}\left[X^{(v)^{\top}} X^{(v)}(I-S)\right]
\end{equation}

\subsubsection{\textbf{Update Multiplier}}

In each iteration, after updating other variables, the multiplier $Y$ can be updated as
\begin{equation}
\label{eq:Yproblem}
Y=Y+\mu(S-C)
\end{equation}

By iteratively updating all variables until convergence or reaching the maximum number of iterations, a unified view-commonness matrix, i.e., the common representation matrix $C$, can be obtained, based on which the affinity matrix can be constructed and the final clustering result can be achieved via the spectral clustering. For clarity, we summarize the overall algorithm of our JSMC approach in Algorithm \ref{algorithm:JSMC}.

\begin{algorithm}[h]
	\caption{Jointly Smoothed Multi-view subspace Clustering (JSMC).}
	\label{algorithm:JSMC}
	\begin{algorithmic}[1]
		
		\STATE \textbf{Input}: Multi-view dataset $X=\{X^{(1)}, X^{(2)}, \ldots, X^{(V)} \}$, the number of clusters $n_c$, the parameters $\alpha$, $\beta$ and $\lambda$.
		\STATE \textbf{Initialization}: $C$ is initialized as the average $K$-NN graph for all the views, $S=0$, $E^{(v)}=0$, $Y=0$.
		
		\REPEAT
		
		\STATE Update variable $S$ by solving Eq.~(\ref{eq:Sproblem}).
		
		\STATE Update view-commonness matrix $C$ by solving Eq.~(\ref{eq:Cproblem}).
		
		\FOR {$ v=1, \ldots, V$}
		
		\STATE Update view-inconsistency matrix $E^{(v)}$ by solving Eq.~(\ref{eq:Eproblem}).
		
		\ENDFOR
		
		\STATE Update the multiplier $Y$ by solving Eq.~(\ref{eq:Yproblem}).
		
		\UNTIL {Convergence or reaching the maximum number of iterations}
		
		\STATE Construct the affinity matrix $W=(|C|+\left|C^{\top}\right|)/2$.
		
		\STATE Partition $W$ into $n_c$ clusters via the spectral clustering .
		
		\STATE \textbf{Output}: The clustering result with $n_c$ clusters.
	\end{algorithmic}
\end{algorithm}

\subsection{Computational Complexity}
\label{sec:complexity}
In this section, we analyze the computational complexity of the proposed JSMC approach (as described in Algorithm \ref{algorithm:JSMC}).

In the initialization of $C$, the construction of the $K$-NN graphs on $V$ views takes $O(n^2V)$ time. In each iteration of JSMC, the first step is to update the auxiliary variable $S$ by solving the standard Sylvester equation via the Bartels-Stewart algorithm, whose time complexity is $O(n^3)$. In the second step, the main computational cost for updating the common representation matrix $C$ comes from the SVD operation, which takes $O(n^3)$ time. Then, updating $E^{(v)}$ with the inverse operation in the third step takes $O(n^3V)$ time. Therefore, the time complexity of one iteration is $O(n^3V)$. After the optimization, the final clustering is obtained by the spectral clustering, which involves the SVD and takes $O(n^3)$ time. Thus the time complexity of the proposed approach is $O(n^3Vt)$, where $t$ is the number of iterations. The space complexity of the proposed approach is $O(n^2)$ due to the storage of several $n\times n$ matrices.

\subsection{Convexity Analysis}
In this section, we analyze the convexity of our objective function.
For the objective function~(\ref{eq:ObjFunction}),  an auxiliary variable $S$ is introduced for the common representation matrix $C$ such that $S=C$. For the $S$-subproblem, the function w.r.t. the auxiliary variable $S$ (in Eq.~(\ref{eq:Sproblem})) is a standard Sylvester equation, with which we can directly obtain the unique solution of $S$. The Hessian matrix of Eq.~(\ref{eq:Sproblem}) is
\begin{equation}
    \begin{aligned}
    &\frac{\partial ^{2} \mathcal{L}}{\partial S \partial S^{\top}} = 2 \sum_{v=1}^V\,(X^{(v)^{\top}} X^{(v)} + \alpha L^{(v)}) + \mu I
    \end{aligned}
\end{equation}
where $I$ is an identity matrix. According to Theorem~\ref{th1}, for any non-zero vector $f$, we have
\begin{equation}
    \begin{aligned}
    f^{\top} \left( 2 \sum_{v=1}^V\,X^{(v)^{\top}} X^{(v)} + \mu I \right)f & = 2 f^{\top} \left(\sum_{v=1}^V\,X^{(v)^{\top}} X^{(v)}\right) f + \mu f^{\top} I f \\
    &=2 \sum_{v=1}^V \| X^{(v)}f \|_F^2 + \mu \| f \|_2^2 \geq 0
    \end{aligned}
\end{equation}
With the Laplacian matrix $L^{(v)}$ being positive semi-definite, the Hessian matrix of Eq.~(\ref{eq:Sproblem}) is also a positive semi-definite matrix. Thus, the subproblem w.r.t. $S$ in Eq.~(\ref{eq:Sproblem}) is convex \cite{wang2020gmc}.

\begin{theorem}
\label{th1}
Given a real symmetric matrix $A \in \mathbb{R}^{n \times n}$, $A$ is a positive semi-definite matrix if for any non-zero vector $f \in \mathbb{R}^n$, $f^{\top} A f \geq 0$ holds.
\end{theorem}

For the $C$-subproblem, we update the view-commonness matrix $C$ while fixing the other variables. The objective function~(\ref{eq:originalC}) w.r.t. $C$ is convex according to \cite{cai2010singular}.

For the $E^{(v)}$-subproblem, we update the view-inconsistency matrix $E^{(v)}$ in objective~(\ref{eq:Eupdate}) while fixing the other variables. Note that each $E^{(v)}$ can be updated respectively. For each $E^{(v)}$, the Hessian matrix can be obtained as
\begin{equation}
    \begin{aligned}
    &\frac{\partial ^{2} \mathcal{L}}{\partial E^{(v)} \partial E^{(v)^{\top}}} = 2 X^{(v)^{\top}} X^{(v)} + 2\beta I
    \end{aligned}
\end{equation}
For any non-zero vector $f$, we have
\begin{equation}
    \begin{aligned}
    f^{\top} \left( 2 X^{(v)^{\top}} X^{(v)} + 2 \beta I \right)f & = 2 f^{\top} X^{(v)^{\top}} X^{(v)} f + 2 \beta f^{\top} I f \\
    &=2 \| X^{(v)}f \|_F^2 + 2 \beta \| f \|_2^2 \geq 0
    \end{aligned}
\end{equation}
Therefore, the Hessian matrix of each $E^{(v)}$ is a positive semi-definite matrix. Thus the subproblem w.r.t. $E^{(v)}$ is convex \cite{wang2020gmc}.

\section{Experiment}
\label{sec:experimet}
In this section, we conduct extensive experiments on a variety of real-world datasets to evaluate the proposed JSMC approach against the state-of-the-art multi-view clustering approaches. The details of datasets and evaluation measures are introduced in Section~\ref{sec:dataset_eval}. The baseline approaches are presented in Section~\ref{sec:baselines}. The  performance comparison of the proposed approach against other multi-view clustering approaches is reported in Section~\ref{sec:cmp_result}. The sensitivity of the parameters is evaluated in Section~\ref{sec:para_analysis}. Finally, the convergence analysis and ablation study are provided in Sections~\ref{sec:convergence} and \ref{sec:ablation}, respectively.

\subsection{Datasets ana Evaluation Measures}
\label{sec:dataset_eval}

In our experiments, eight real-world datasets are used, including \emph{3Sources}, \emph{Notting-Hill}, \emph{ORL}, \emph{WebKB-Texas}, \emph{Yale}, \emph{Reuters}, \emph{COIL-20}, and \emph{Caltech-7}, which will be introduced as follows.

\begin{itemize}
\item \textbf{\emph{3Sources}.}
This dataset is a multi-view text dataset with $948$ news articles, which includes three online news sources, i.e., BBC, Reuters, and Guardian~\cite{wang2019auto}. In the \emph{3Sources} dataset, each news article belongs to one of the six categories, i.e., business, politics, health, entertainment, sport, and technology. In the experiments, only data instances that are complete in all views are selected. The dimensions of the three views, i.e., BBC, Reuters and Guardian, are  $3068$, $3631$, and $3560$, respectively.

\item \textbf{\emph{Notting-Hill}.}
This dataset is a video image dataset from the movie ``Notting Hill"~\cite{zhang2009character}. In $76$ video clips from the movie, $4660$ face images of $5$ characters are collected, where each face image has a pixel-size of $120\times150$. In the experiments, $550$ face images are included, which are associated with three views, namely, Intensity (2000-D), Gabor (6750-D), and LBP (3304-D).

\item \textbf{\emph{ORL}.}
This dataset consists of $400$ face images from $40$ persons \cite{hu20_if}. Each face image is taken with different facial expressions, facial details, and lighting conditions. In the \emph{ORL} dataset, three views are included, namely, Intensity (4096-D),  Gabor (6750-D), and LBP (3304-D).

\item \textbf{\emph{WebKB-Texas}.}
This dataset is a collection of 187 documents from the homepage of the Computer Science Department of the University of Texas~\cite{craven1998learning}. These documents are grouped into the following tags: student, project, course, staff and faculty. Each document is associated with two views, i.e., citation and content, whose dimensions are $187$ and $1703$, respectively.

\item \textbf{\emph{Yale}.}
This dataset consists of $165$ face images from $15$ persons \cite{chen21_if}, each of which has $11$ gray-scale images with different facial expressions, such as with glasses, without glasses, center-light, left-light, right-light, normal, happy, sad, sleepy, surprised, and wink. In this dataset, three views are included, namely, Intensity (4096-D),  Gabor (6750-D), and LBP (3304-D).

\item \textbf{\emph{Reuters}.}
This dataset consists of $1200$ documents \cite{zhang2018generalized}, which are divided into six categories, namely, E$21$, CCAT, M$11$, GCAT, C$15$, and ECAT. Each document in the \emph{Reuters} dataset is associated with five views, corresponding to  five different languages, i.e., English (2000-D), French (2000-D), German (2000-D), Italian (2000-D), and Spanish (2000-D).

\item \textbf{\emph{COIL-20}.}
This dataset is a collection of gray-scale images of $20$ objects~\cite{nene1996columbia}, which are taken from different angles with $72$ poses each. In the \emph{COIL-20} dataset, there are three different views, i.e., Intensity (1024-D), LBP (3304-D), and Gabor (6750-D).

\item \textbf{\emph{Caltech-7}.}
This dataset is a subset of the Caltech-101 dataset~\cite{fei2004learning}, which has images of objects belonging to $101$ classes. Each image is about $200\times300$ pixels in this subset. In the experiments, we select images from $7$ classes, including Faces, Garfield, Stop-Sign, Snoopy, Windsor-Chair, Motorbikes, and Dollar Bill.  Three views are extracted for each image, namely, GIST (512-D), HOG (1984-D), and LBP (928-D).
\end{itemize}

To evaluate the performances of different multi-view clustering algorithms, four widely-used evaluation measures are adopted, namely, normalized mutual information (NMI)~\cite{strehl02},  adjusted Rand index (ARI) \cite{Huang2021}, accuracy (ACC) \cite{liu19_pami}, and purity (PUR) \cite{zhang2018binary}.

\subsection{Baseline Methods and Experimental Settings}
\label{sec:baselines}

In the experiments, we compare the proposed JSMC method against the spectral clustering method as well as seven multi-view clustering methods, which will be described as follows.

\begin{itemize}
\item {\textbf{SC$_{best}$} \cite{ng2002spectral}.}
Spectral clustering (SC) is a classical (single-view) clustering method. In the experiments, we conduct spectral clustering on each single view, and report its best single-view performance.

	\item {\textbf{MVSpec}~\cite{Tzor12_icdm}.}
	Multi-view spectral clustering (MVSpec) represents each view with a kernel matrix and captures the importance of different views via a weighted combination of kernels for multi-view clustering.

\item {\textbf{RMSC}~\cite{xia2014robust}.}
Robust multi-view spectral clustering (RMSC) constructs a unified probability matrix from multiple probability matrices with the low-rank constraint. Then the unified matrix is partitioned for obtaining the final clustering.

\item {\textbf{MVSC}~\cite{li2015large}.}
Multi-view spectral clustering (MVSC) first constructs a bipartite graph between the data instances and a set of anchors on each view, and then fuses the multiple bipartite graphs from different views into a unified bipartite graph, upon which the graph partitioning is performed to obtain the clustering result.

\item {\textbf{MVGL}~\cite{zhan2017graph}.}
Multi-view graph learning (MVGL) fuses multiple affinity graphs from multiple views into a unified graph with the rank constraint, and obtains the clustering result by partitioning the unified graph.

\item {\textbf{BMVC}~\cite{zhang2018binary}.}
Binary multi-view clustering (BMVC) solves the multi-view clustering problem by binary representation, which simultaneously optimizes the binary learning and the clustering for the multi-view dataset.

\item {\textbf{SMVSC}~\cite{sun2021scalable}.}
Scalable multi-view subspace clustering (SMVSC) jointly optimizes the anchor learning and the unified graph construction, where the latent structures of data can be expressed by the subspace representation via a set of unified anchors.

\item {\textbf{FPMVS-CAG}~\cite{wang2021fast}.}
Fast parameter-free multi-view subspace clustering with consensus anchor guidance (FPMVS-CAG) combines the anchor selection and the graph construction into a parameter-free manner, and learns an anchor-based subspace representation for the final clustering.

\end{itemize}

In the experiments, we adopt the grid search scheme to find best parameters on each dataset. For the proposed method and all the baseline methods, each parameter is searched in the range of  $\{10^{-5},10^{-4},\cdots,10^{5}\}$, unless a specific range of the parameter is suggested in the corresponding paper.

\begin{table}[!t]
	\renewcommand\arraystretch{1.25}
	\caption{Average performances (w.r.t. NMI(\%)) of different methods. The best score in each row is highlighted in bold.}\vskip 0.02 in
	\label{tab:nmi}
	\scalebox{0.65}{
		\centering
		\begin{tabular}{m{2.25cm}<{\centering}|m{1.47cm}<{\centering}m{1.47cm}<{\centering}m{1.53cm}<{\centering}m{1.53cm}<{\centering}m{1.52cm}<{\centering}m{1.47cm}<{\centering}m{1.47cm}<{\centering}m{1.52cm}<{\centering}m{1.64cm}<{\centering}}
			\toprule
			Method    &SC$_{best}$   &MVSpec &RMSC   &MVSC   &MVGL   &BMVC   &SMVSC   &FPMVS-CAG   &JSMC\\
			\midrule
			\emph{3Sources}		&38.69$_{\pm0.00}$	&6.61$_{\pm0.00}$	&58.56$_{\pm3.50}$	&66.48$_{\pm3.46}$	&8.07$_{\pm0.00}$	&58.69$_{\pm0.00}$	&8.92$_{\pm0.00}$	&6.59$_{\pm0.00}$	&\textbf{69.52}$_{\pm0.00}$\\
			\emph{Notting-Hill}		&72.62$_{\pm0.00}$	&2.22$_{\pm0.00}$	&71.29$_{\pm3.68}$	&77.09$_{\pm8.11}$	&85.51$_{\pm0.00}$	&4.14$_{\pm0.00}$	&82.49$_{\pm0.00}$	&71.35$_{\pm0.00}$	&\textbf{96.92}$_{\pm0.00}$\\
			\emph{ORL}		&89.12$_{\pm0.00}$	&39.42$_{\pm0.00}$	&84.79$_{\pm1.43}$	&84.69$_{\pm2.06}$	&82.96$_{\pm0.00}$	&39.29$_{\pm0.00}$	&75.26$_{\pm0.00}$	&74.30$_{\pm0.00}$	&\textbf{91.46}$_{\pm0.00}$\\
			\emph{WebKB-Texas}		&29.36$_{\pm0.00}$	&10.15$_{\pm0.00}$	&26.01$_{\pm5.69}$	&1.98$_{\pm0.25}$	&6.55$_{\pm0.00}$	&24.73$_{\pm0.00}$	&21.42$_{\pm0.00}$	&22.76$_{\pm0.00}$	&\textbf{38.04}$_{\pm0.00}$\\
			\emph{Yale}		&64.94$_{\pm0.00}$	&25.41$_{\pm0.00}$	&66.36$_{\pm2.76}$	&64.10$_{\pm3.25}$	&67.15$_{\pm0.00}$	&27.57$_{\pm0.00}$	&61.68$_{\pm0.00}$	&49.76$_{\pm0.00}$	&\textbf{76.38}$_{\pm0.00}$\\
			\emph{Reuters}		&23.67$_{\pm0.00}$	&1.85$_{\pm0.00}$	&\textbf{33.01}$_{\pm1.68}$	&1.26$_{\pm0.46}$	&4.31$_{\pm0.00}$	&22.46$_{\pm0.00}$	&18.87$_{\pm0.00}$	&20.64$_{\pm0.00}$	&25.96$_{\pm0.00}$\\
			\emph{COIL-20}		&80.80$_{\pm0.00}$	&24.36$_{\pm0.00}$	&80.42$_{\pm1.75}$	&75.68$_{\pm4.33}$	&91.80$_{\pm0.00}$	&50.69$_{\pm0.00}$	&73.48$_{\pm0.00}$	&74.63$_{\pm0.00}$	&\textbf{92.98}$_{\pm0.00}$\\
			\emph{Caltech-7}		&38.26$_{\pm0.00}$	&10.79$_{\pm0.00}$	&42.11$_{\pm1.33}$	&54.46$_{\pm11.37}$	&55.98$_{\pm0.00}$	&47.29$_{\pm0.00}$	&47.79$_{\pm0.00}$	&47.32$_{\pm0.00}$	&\textbf{63.20}$_{\pm0.00}$\\
			\midrule
			Avg. score		&54.68	&15.10	&57.82	&53.22	&50.29	&34.36	&48.74	&45.92	&\textbf{69.31}\\
			Avg. rank		&4.00	&8.38   &4.00	&4.88	&4.25	&6.13	&5.50	&6.13	&\textbf{1.13}\\
			\bottomrule
		\end{tabular}
	}
\end{table}

\begin{table}[!t]
	\renewcommand\arraystretch{1.25}
	\caption{Average performances (w.r.t. ARI(\%)) of different methods. The best score in each row is highlighted in bold.}\vskip 0.02 in
	\label{tab:ari}
	\scalebox{0.65}{
		\centering
		\begin{tabular}{m{2.25cm}<{\centering}|m{1.47cm}<{\centering}m{1.50cm}<{\centering}m{1.53cm}<{\centering}m{1.53cm}<{\centering}m{1.52cm}<{\centering}m{1.47cm}<{\centering}m{1.47cm}<{\centering}m{1.52cm}<{\centering}m{1.64cm}<{\centering}}
			\toprule
			Method    &SC$_{best}$  &MVSpec  &RMSC   &MVSC   &MVGL   &BMVC   &SMVSC   &FPMVS-CAG   &JSMC\\
			\midrule
			\emph{3Sources}		&20.81$_{\pm0.00}$	&0.85$_{\pm0.00}$	&48.87$_{\pm3.39}$	&58.38$_{\pm4.37}$	&-0.72$_{\pm0.00}$	&54.32$_{\pm0.00}$	&4.00$_{\pm0.00}$	&1.01$_{\pm0.00}$	&\textbf{65.17}$_{\pm0.00}$\\
			\emph{Notting-Hill}		&73.93$_{\pm0.00}$	&1.09$_{\pm0.00}$	&69.26$_{\pm5.15}$	&74.52$_{\pm9.88}$	&81.74$_{\pm0.00}$	&2.66$_{\pm0.00}$	&83.95$_{\pm0.00}$	&62.25$_{\pm0.00}$	&\textbf{97.69}$_{\pm0.00}$\\
			\emph{ORL}		&69.75$_{\pm0.00}$	&1.98$_{\pm0.00}$	&61.42$_{\pm3.28}$	&56.95$_{\pm6.96}$	&38.79$_{\pm0.00}$	&0.61$_{\pm0.00}$	&42.39$_{\pm0.00}$	&39.99$_{\pm0.00}$	&\textbf{77.79}$_{\pm0.00}$\\
			\emph{WebKB-Texas}		&23.91$_{\pm0.00}$	&-4.64$_{\pm0.00}$	&18.51$_{\pm5.09}$	&0.19$_{\pm0.35}$	&2.48$_{\pm0.00}$	&25.18$_{\pm0.00}$	&29.01$_{\pm0.00}$	&29.81$_{\pm0.00}$	&\textbf{32.26}$_{\pm0.00}$\\
			\emph{Yale}		&45.06$_{\pm0.00}$	&0.26$_{\pm0.00}$	&45.64$_{\pm3.98}$	&40.15$_{\pm5.28}$	&41.53$_{\pm0.00}$	&1.77$_{\pm0.00}$	&37.56$_{\pm0.00}$	&25.30$_{\pm0.00}$	&\textbf{57.58}$_{\pm0.00}$\\
			\emph{Reuters}		&17.19$_{\pm0.00}$	&0.24$_{\pm0.00}$	&\textbf{26.15}$_{\pm1.15}$	&0.05$_{\pm0.03}$	&0.27$_{\pm0.00}$	&17.00$_{\pm0.00}$	&11.85$_{\pm0.00}$	&16.91$_{\pm0.00}$	&20.48$_{\pm0.00}$\\
			\emph{COIL-20}		&66.79$_{\pm0.00}$	&4.48$_{\pm0.00}$	&66.14$_{\pm3.19}$	&51.95$_{\pm8.71}$	&78.21$_{\pm0.00}$	&27.83$_{\pm0.00}$	&51.14$_{\pm0.00}$	&53.21$_{\pm0.00}$	&\textbf{81.28}$_{\pm0.00}$\\
			\emph{Caltech-7}		&30.55$_{\pm0.00}$	&-1.73$_{\pm0.00}$	&32.05$_{\pm2.62}$	&47.92$_{\pm14.14}$	&41.99$_{\pm0.00}$	&36.34$_{\pm0.00}$	&38.21$_{\pm0.00}$	&41.34$_{\pm0.00}$	&\textbf{60.10}$_{\pm0.00}$\\
			\midrule
			Avg. score		&43.50	&0.32	&46.00	&41.26	&35.54	&20.71	&37.26	&33.73	&\textbf{61.54}\\
			Avg. rank		&4.25	&8.63   &4.13	&4.88	&5.13	&6.13	&5.00	&5.38	&\textbf{1.13}\\
			\bottomrule
		\end{tabular}
	}
\end{table}

\subsection{Performance Comparison and Analysis}
\label{sec:cmp_result}
In this section, we compare our proposed JSMC method with other multi-view clustering methods. Specifically, we run each of the test methods twenty times, and report their average performances w.r.t. to NMI, ARI, ACC, and PUR in Tables~\ref{tab:nmi},~\ref{tab:ari},~\ref{tab:acc}, and \ref{tab:purity}, respectively.

As shown in Table~\ref{tab:nmi}, our JSMC method achieves the best performance w.r.t. NMI on seven out of eight benchmark datasets. Though the RMSC method obtains a higher NMI score than JSMC on the \emph{Reuters} dataset, yet our JSMC method yield higher or significantly higher NMI scores than RMSC on all the other seven datasets. Especially, on the \emph{Notting-Hill} datasets, our JSMC method achieves an NMI(\%) score of 96.92, whereas the second best method (i.e., MVGL) only achieves an NMI(\%) score of 85.51.
Further, we also report the average score and average rank (across the eight benchmark datasets) for the proposed method and the baseline methods. In terms of average score, our JSMC method achieves an average NMI(\%) score (across eight datasets) of 69.31, which is significantly higher than the second best average NMI(\%) score of 57.82 (achieved by RMSC). In terms of average rank, our JSMC method achieves an average rank of 1.13, whereas the second best method only achieves an average rank  of 4.00.

\begin{table}[!t]
	\renewcommand\arraystretch{1.25}
	\caption{Average performances (w.r.t. ACC(\%)) of different methods. The best score in each row is highlighted in bold.}\vskip 0.02 in
	\label{tab:acc}
	\scalebox{0.65}{
		\centering
		\begin{tabular}{m{2.25cm}<{\centering}|m{1.47cm}<{\centering}m{1.47cm}<{\centering}m{1.53cm}<{\centering}m{1.53cm}<{\centering}m{1.52cm}<{\centering}m{1.47cm}<{\centering}m{1.47cm}<{\centering}m{1.52cm}<{\centering}m{1.64cm}<{\centering}}
			\toprule
			Method    &SC$_{best}$  &MVSpec  &RMSC   &MVSC   &MVGL   &BMVC   &SMVSC   &FPMVS-CAG   &JSMC\\
			\midrule
			\emph{3Sources}		&48.52$_{\pm0.00}$	&25.44$_{\pm0.00}$	&62.54$_{\pm4.31}$	&75.86$_{\pm4.58}$	&35.50$_{\pm0.00}$	&65.68$_{\pm0.00}$	&31.36$_{\pm0.00}$	&26.04$_{\pm0.00}$	&\textbf{77.51}$_{\pm0.00}$\\
			\emph{Notting-Hill}		&86.55$_{\pm0.00}$	&26.36$_{\pm0.00}$	&79.41$_{\pm3.22}$	&81.20$_{\pm5.86}$	&84.36$_{\pm0.00}$	&27.27$_{\pm0.00}$	&91.45$_{\pm0.00}$	&70.55$_{\pm0.00}$	&\textbf{98.91}$_{\pm0.00}$\\
			\emph{ORL}		&77.25$_{\pm0.00}$	&19.50$_{\pm0.00}$	&70.23$_{\pm2.73}$	&72.36$_{\pm3.81}$	&72.00$_{\pm0.00}$	&16.75$_{\pm0.00}$	&56.25$_{\pm0.00}$	&56.00$_{\pm0.00}$	&\textbf{82.00}$_{\pm0.00}$\\
			\emph{WebKB-Texas}		&57.75$_{\pm0.00}$	&32.09$_{\pm0.00}$	&48.64$_{\pm4.80}$	&55.13$_{\pm0.30}$	&51.87$_{\pm0.00}$	&56.68$_{\pm0.00}$	&56.68$_{\pm0.00}$	&57.75$_{\pm0.00}$	&\textbf{59.89}$_{\pm0.00}$\\
			\emph{Yale}		&63.64$_{\pm0.00}$	&21.82$_{\pm0.00}$	&61.21$_{\pm4.24}$	&59.03$_{\pm4.26}$	&64.85$_{\pm0.00}$	&22.42$_{\pm0.00}$	&56.97$_{\pm0.00}$	&44.24$_{\pm0.00}$	&\textbf{73.33}$_{\pm0.00}$\\
			\emph{Reuters}		&42.00$_{\pm0.00}$	&20.75$_{\pm0.00}$	&{52.60}$_{\pm2.60}$	&17.59$_{\pm0.43}$	&18.92$_{\pm0.00}$	&45.00$_{\pm0.00}$	&38.08$_{\pm0.00}$	&44.33$_{\pm0.00}$	&46.75$_{\pm0.00}$\\
			\emph{COIL-20}		&72.36$_{\pm0.00}$	&15.76$_{\pm0.00}$	&70.84$_{\pm4.09}$	&61.66$_{\pm6.58}$	&81.39$_{\pm0.02}$	&40.63$_{\pm0.00}$	&60.56$_{\pm0.00}$	&63.75$_{\pm0.00}$	&\textbf{83.96}$_{\pm0.00}$\\
			\emph{Caltech-7}		&46.40$_{\pm0.00}$	&38.87$_{\pm0.00}$	&40.39$_{\pm2.82}$	&63.68$_{\pm6.14}$	&66.28$_{\pm0.00}$	&47.15$_{\pm0.00}$	&46.68$_{\pm0.00}$	&54.55$_{\pm0.00}$	&\textbf{70.90}$_{\pm0.00}$\\
			\hline
			Avg.score		&61.81	&25.07	&60.73	&60.81	&59.40	&40.20	&54.75	&52.15	&\textbf{74.16}\\
			\hline
			Avg. rank		&3.75	&8.63	&5.00	&4.75	&4.25	&5.88	&5.50	&5.50	&\textbf{1.13}\\
			\bottomrule
		\end{tabular}
	}
\end{table}

\begin{table}[!t]
	\renewcommand\arraystretch{1.25}
	\caption{Average performances (w.r.t. PUR(\%)) of different methods. The best score in each row is highlighted in bold.}\vskip 0.02 in
	\label{tab:purity}
	\scalebox{0.65}{
		\centering
		\begin{tabular}{m{2.25cm}<{\centering}|m{1.47cm}<{\centering}m{1.47cm}<{\centering}m{1.53cm}<{\centering}m{1.53cm}<{\centering}m{1.52cm}<{\centering}m{1.47cm}<{\centering}m{1.47cm}<{\centering}m{1.52cm}<{\centering}m{1.64cm}<{\centering}}
			\toprule
			Method    &SC$_{best}$  &MVSpec  &RMSC   &MVSC   &MVGL   &BMVC   &SMVSC   &FPMVS-CAG   &JSMC\\
			\midrule
			\emph{3Sources}		&59.17$_{\pm0.00}$	&39.64$_{\pm0.00}$	&76.33$_{\pm3.60}$	&80.89$_{\pm3.26}$	&40.24$_{\pm0.00}$	&73.37$_{\pm0.00}$	&44.38$_{\pm0.00}$	&42.60$_{\pm0.00}$	&\textbf{82.25}$_{\pm0.00}$\\
			\emph{Notting-Hill}		&86.55$_{\pm0.00}$	&32.36$_{\pm0.00}$	&82.23$_{\pm1.94}$	&83.55$_{\pm6.19}$	&87.09$_{\pm0.00}$	&35.64$_{\pm0.00}$	&91.45$_{\pm0.00}$	&82.73$_{\pm0.00}$	&\textbf{98.91}$_{\pm0.00}$\\
			\emph{ORL}		&80.00$_{\pm0.00}$	&21.25$_{\pm0.00}$	&74.31$_{\pm2.22}$	&77.05$_{\pm2.47}$	&77.75$_{\pm0.00}$	&18.00$_{\pm0.00}$	&60.25$_{\pm0.00}$	&60.00$_{\pm0.00}$	&\textbf{84.50}$_{\pm0.00}$\\
			\emph{WebKB-Texas}		&69.52$_{\pm0.00}$	&55.08$_{\pm0.00}$	&68.80$_{\pm4.35}$	&56.10$_{\pm0.16}$	&57.22$_{\pm0.00}$	&68.98$_{\pm0.00}$	&67.91$_{\pm0.00}$	&65.78$_{\pm0.00}$	&\textbf{70.05}$_{\pm0.00}$\\
			\emph{Yale}		&64.24$_{\pm0.00}$	&22.42$_{\pm0.00}$	&62.36$_{\pm4.02}$	&60.33$_{\pm3.59}$	&64.85$_{\pm0.00}$	&24.24$_{\pm0.00}$	&56.97$_{\pm0.00}$	&46.67$_{\pm0.00}$	&\textbf{73.94}$_{\pm0.00}$\\
			\emph{Reuters}		&43.75$_{\pm0.00}$	&20.83$_{\pm0.00}$	&\textbf{54.40}$_{\pm1.81}$	&17.90$_{\pm0.45}$	&20.83$_{\pm0.00}$	&45.33$_{\pm0.00}$	&40.50$_{\pm0.00}$	&46.83$_{\pm0.00}$	&48.58$_{\pm0.00}$\\
			\emph{COIL-20}		&74.72$_{\pm0.00}$	&23.19$_{\pm0.00}$	&72.51$_{\pm3.11}$	&64.69$_{\pm5.64}$	&86.25$_{\pm0.00}$	&40.76$_{\pm0.00}$	&61.67$_{\pm0.00}$	&65.42$_{\pm0.00}$	&\textbf{88.68}$_{\pm0.00}$\\
			\emph{Caltech-7}		&82.29$_{\pm0.00}$	&55.16$_{\pm0.00}$	&84.13$_{\pm1.35}$	&82.91$_{\pm7.85}$	&85.35$_{\pm0.00}$	&84.94$_{\pm0.00}$	&87.31$_{\pm0.00}$	&84.74$_{\pm0.00}$	&\textbf{92.06}$_{\pm0.00}$\\
			\hline
			Avg.score		&70.03	&33.74	&71.88	&65.43	&64.95	&48.91	&63.81	&61.85	&\textbf{79.87}\\
			\hline
			Avg. rank		&4.00	&8.63	&4.25	&5.75	&4.38	&6.00	&5.00	&5.75	&\textbf{1.13}\\
			\bottomrule
		\end{tabular}
	}
\end{table}

\begin{figure*}[!t]
	\begin{center}
		\includegraphics[width=1\textwidth]{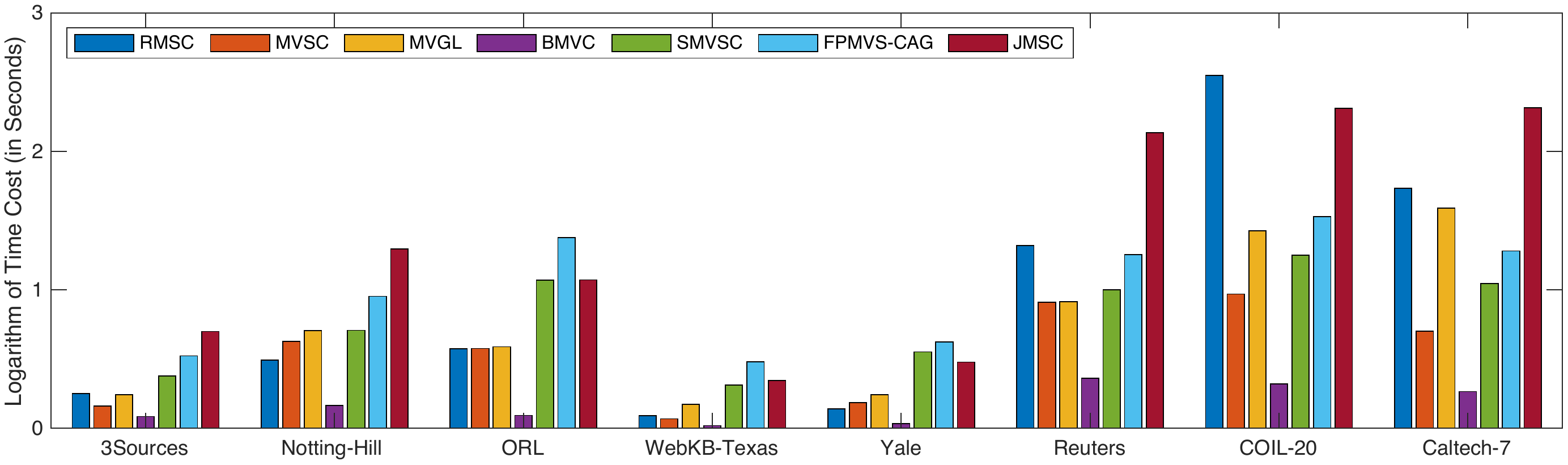}
		\caption{Running time of different algorithms on the eight datasets.}\vskip -0.15 in
		\label{fig:timecost}
	\end{center}
\end{figure*}

Similar advantages of the proposed JSMC method can also be observed in Tables~\ref{tab:ari},~\ref{tab:acc}, and \ref{tab:purity}, corresponding to the ARI, ACC, and PUR scores, respectively. In terms of ARI(\%), our JSMC method achieves an average score of 61.54 and an average rank of 1.13, whereas the second best method achieves an average score of 46.00 and an average rank of 4.13. In terms of ACC(\%), our JSMC method achieves an average score of 74.16 and an average rank of 1.13, whereas the second best method achieves an average score of 61.81 and an average rank of 3.75. In terms of PUR(\%), our JSMC method achieves an average score of 79.87 and an average rank of 1.13, whereas the second best method achieves an average score of 71.88 and an average rank of 3.75. To summarize, the experimental results in Tables~\ref{tab:nmi},~\ref{tab:ari},~\ref{tab:acc}, and \ref{tab:purity} have shown the accuracy and robustness of the proposed JSMC method over the other multi-view clustering methods.

\subsection{Analysis of Time and Memory Costs}
\label{sec:time_space_cost}

\begin{figure*}[!t]
	\begin{center}
		\includegraphics[width=0.56\textwidth]{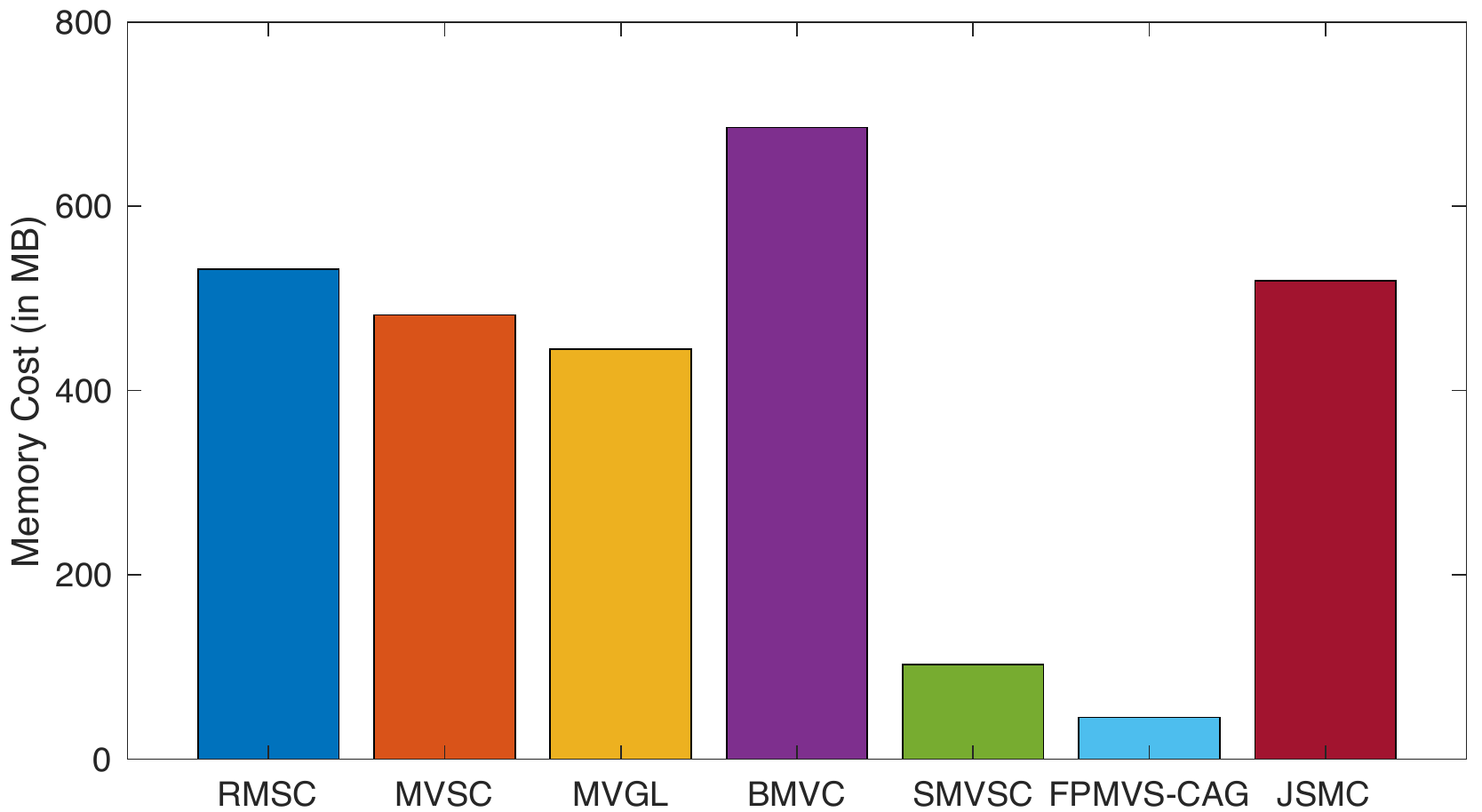}
		\caption{Memory costs of different algorithms on the \emph{Caltech-7} dataset.}
		\label{fig:memorycost}
	\end{center}
\end{figure*}

In this section, we compare the time and memory costs of different methods in Figs.~\ref{fig:timecost} and \ref{fig:memorycost}, respectively.
In terms of time cost, as shown in Fig.~\ref{fig:timecost}, our JSMC method is faster than FPMVS-CAG on the \emph{ORL}, \emph{WebKB-Texas}, and \emph{Yale} datasets, and slower than FPMVS-CAG on the other datasets. In terms of the memory cost, we compare the memory costs of different methods on the \emph{Caltech-7} dataset.
As shown in Fig.~\ref{fig:memorycost}, the two bipartite graph based methods (i.e., SMVSC and FPMVS) consume less memory than the other methods on the \emph{Caltech-7} dataset, while the memory cost of our JSMC method is comparable  to RMSC, MVSC, MVGL, and BMVC.
To summarize, as can be observed in Tables Tables~\ref{tab:nmi}, \ref{tab:ari}, \ref{tab:acc}, and \ref{tab:purity} and Figs.~\ref{fig:timecost} and \ref{fig:memorycost}, our JSMC method requires comparable computational costs to some general baseline methods, while producing highly-competitive (or advantageous) clustering performance on the benchmark datasets.

\subsection{Parameter Analysis}
\label{sec:para_analysis}
\begin{figure*}[!t]
	\begin{center}
		{\subfigure[\emph{3Sources}]
			{\includegraphics[width=0.23\columnwidth]{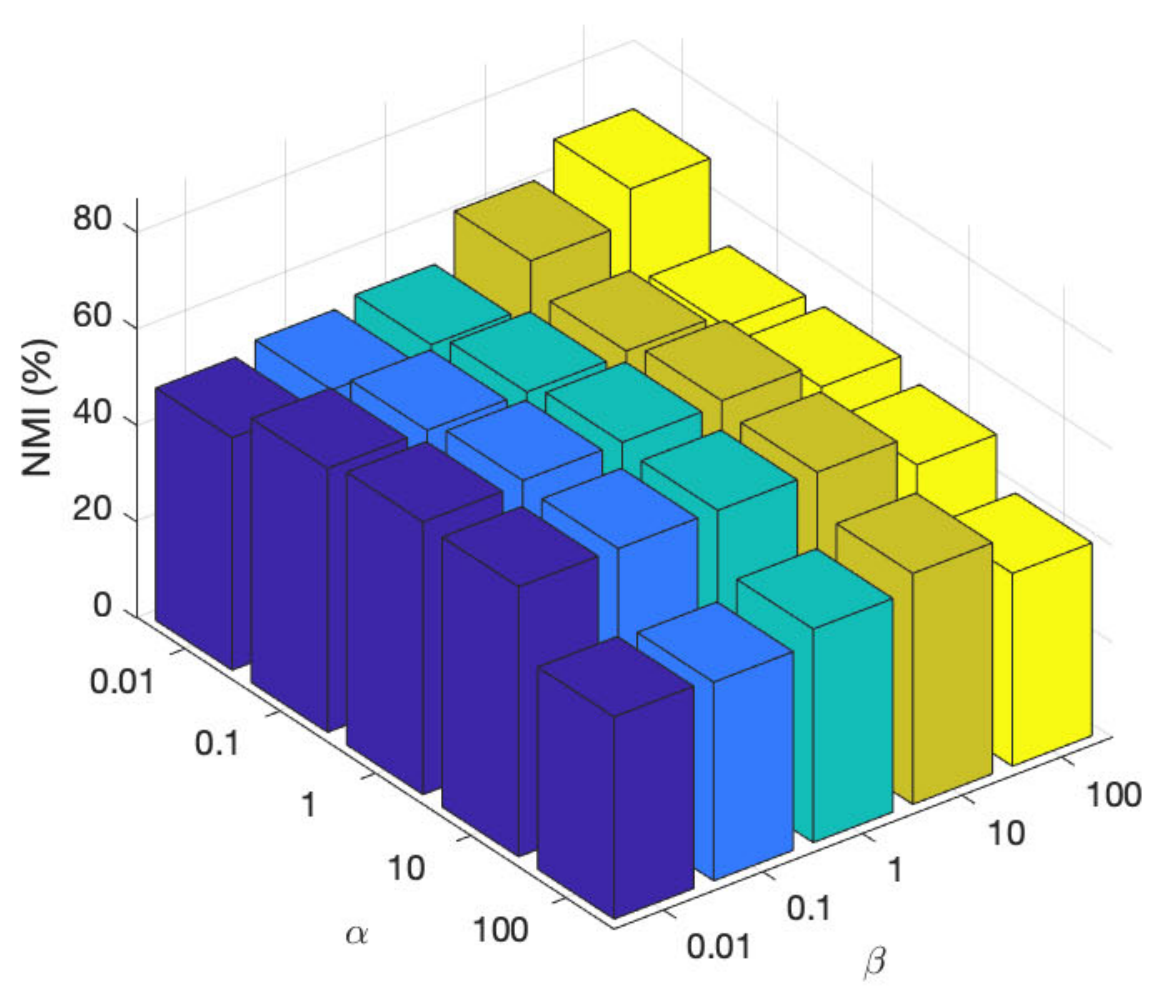}}}
		{\subfigure[\emph{Notting-Hill}]
			{\includegraphics[width=0.23\columnwidth]{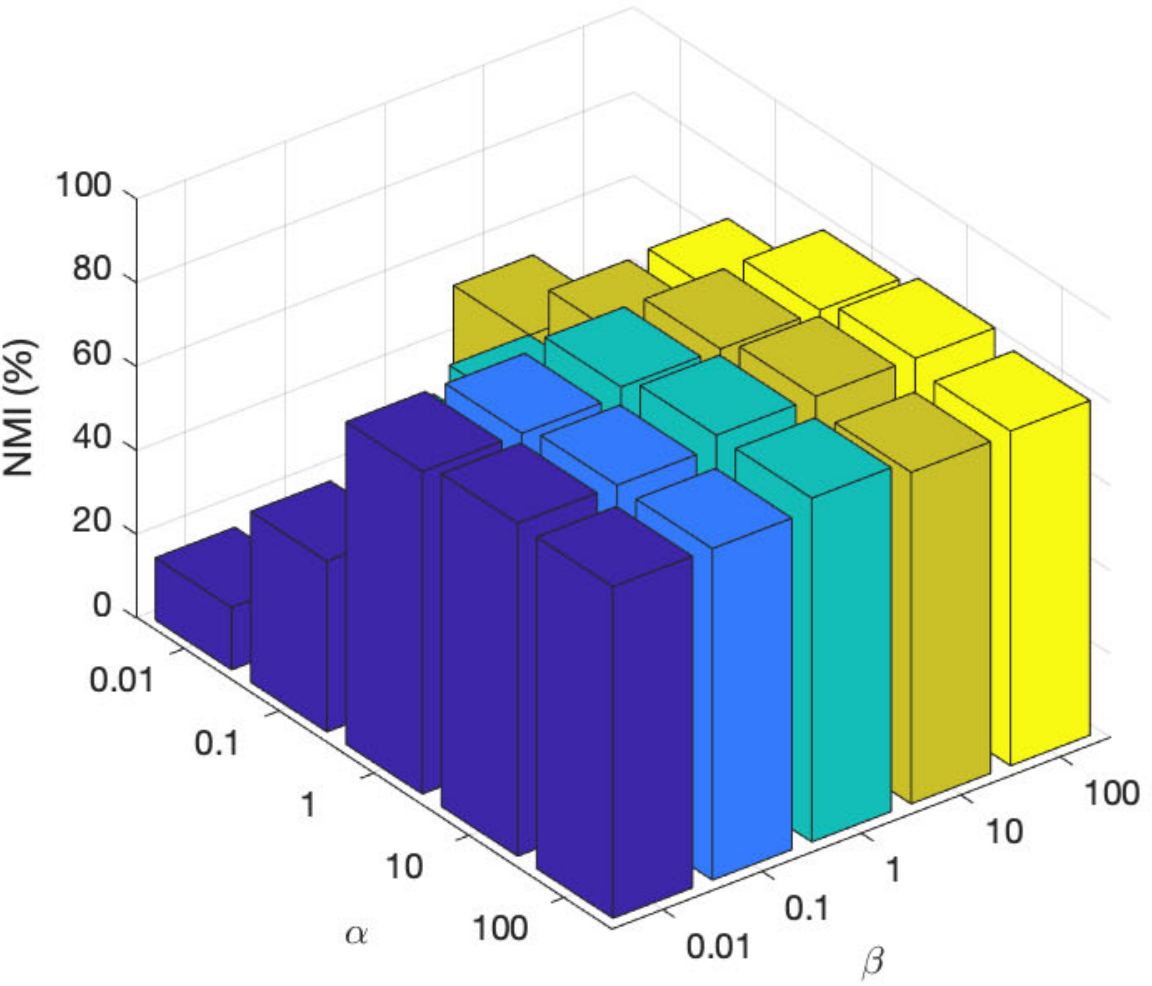}}}
		{\subfigure[\emph{ORL}]
			{\includegraphics[width=0.23\columnwidth]{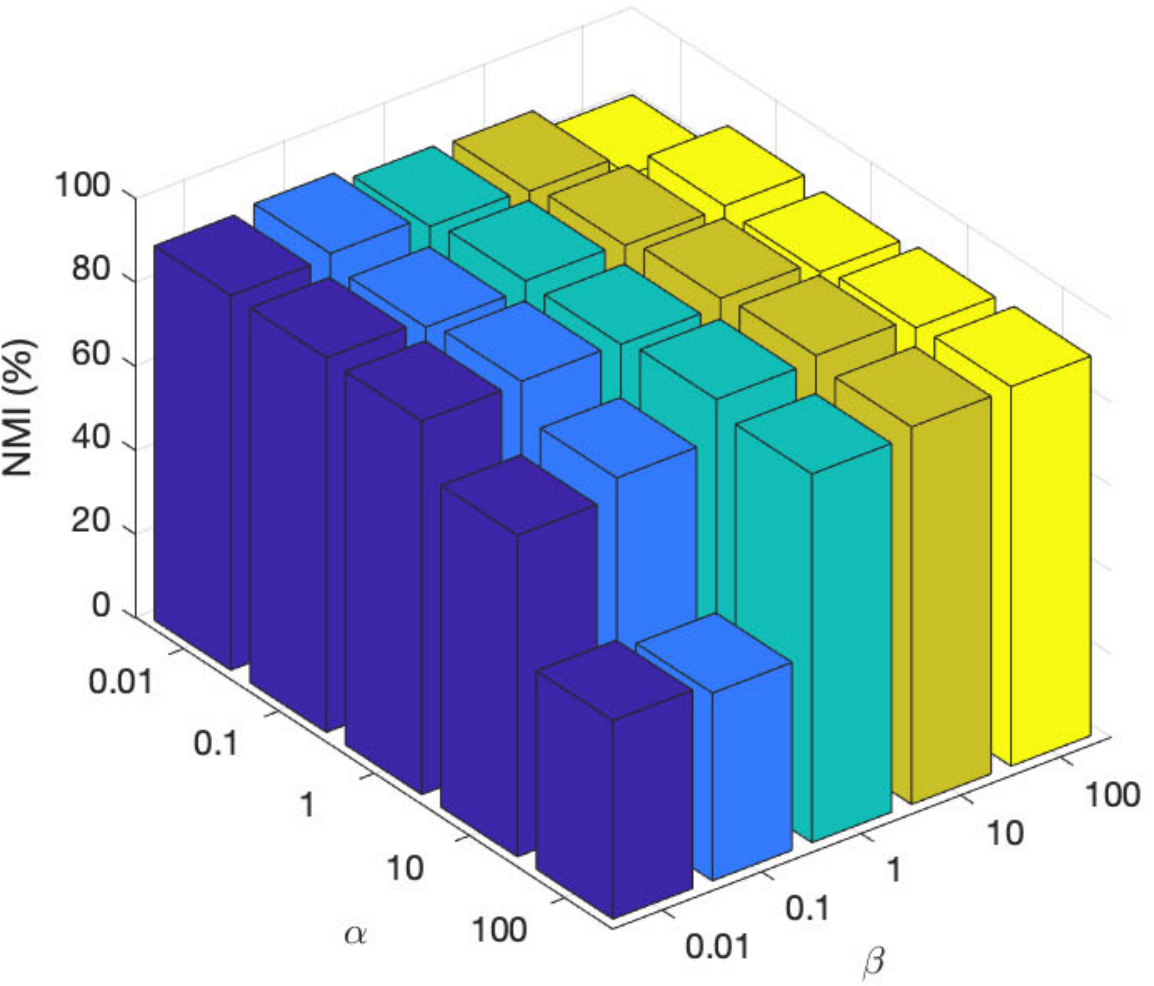}}}
		{\subfigure[\emph{WebKB-Texas}]
			{\includegraphics[width=0.23\columnwidth]{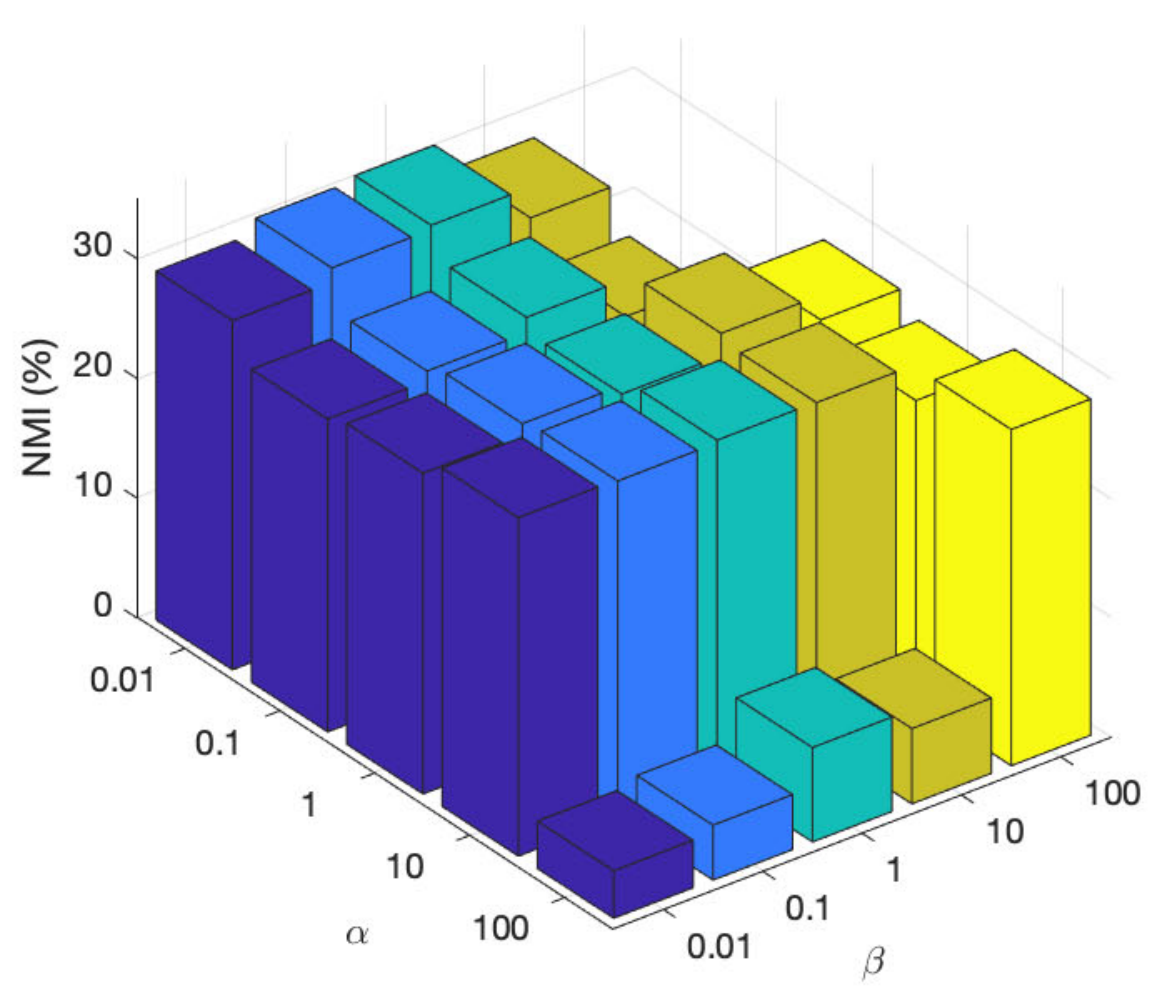}}}
		{\subfigure[\emph{Yale}]
			{\includegraphics[width=0.23\columnwidth]{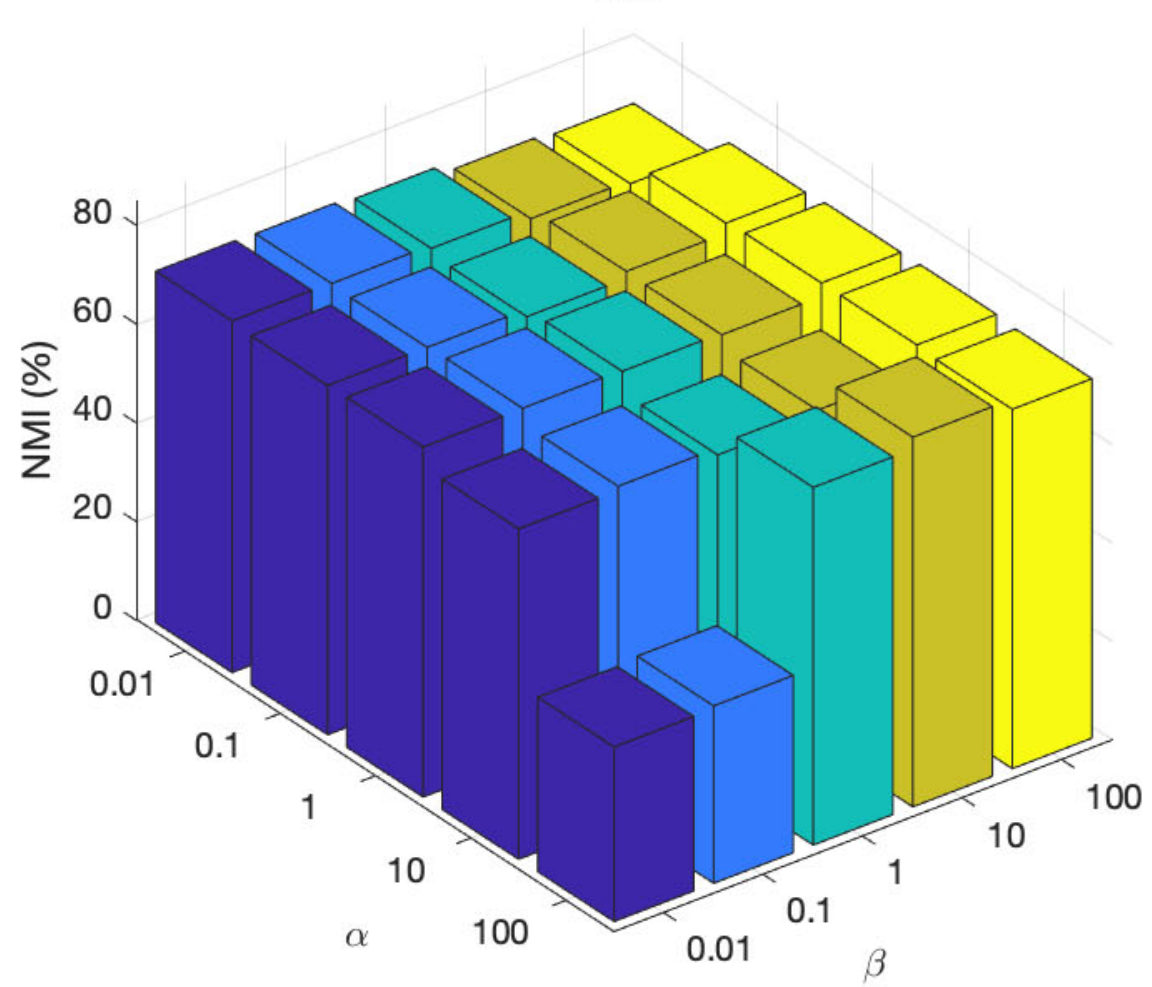}}}
		{\subfigure[\emph{Reuters}]
			{\includegraphics[width=0.23\columnwidth]{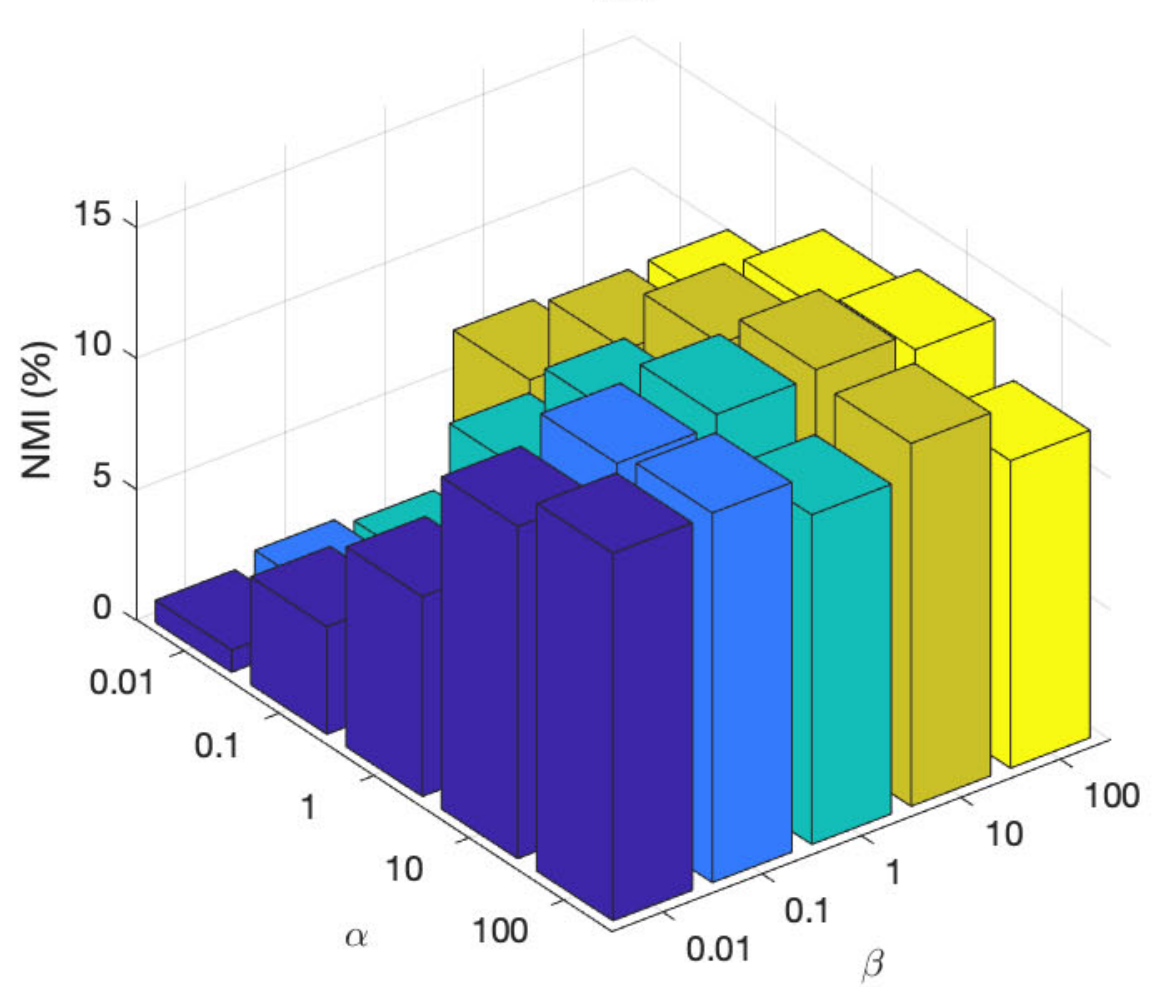}}}
		{\subfigure[\emph{COIL-20}]
			{\includegraphics[width=0.23\columnwidth]{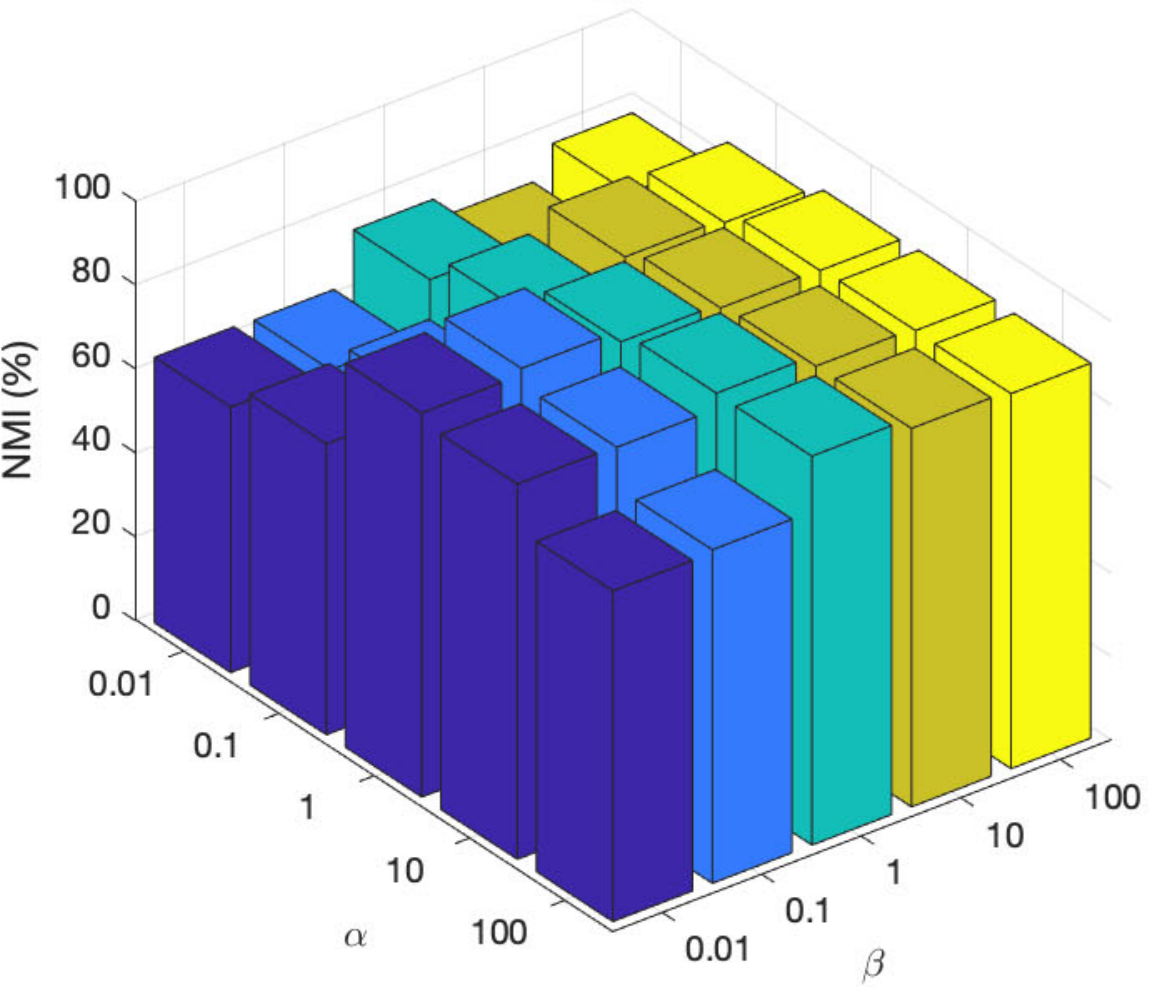}}}
		{\subfigure[\emph{Caltech-7}]
			{\includegraphics[width=0.23\columnwidth]{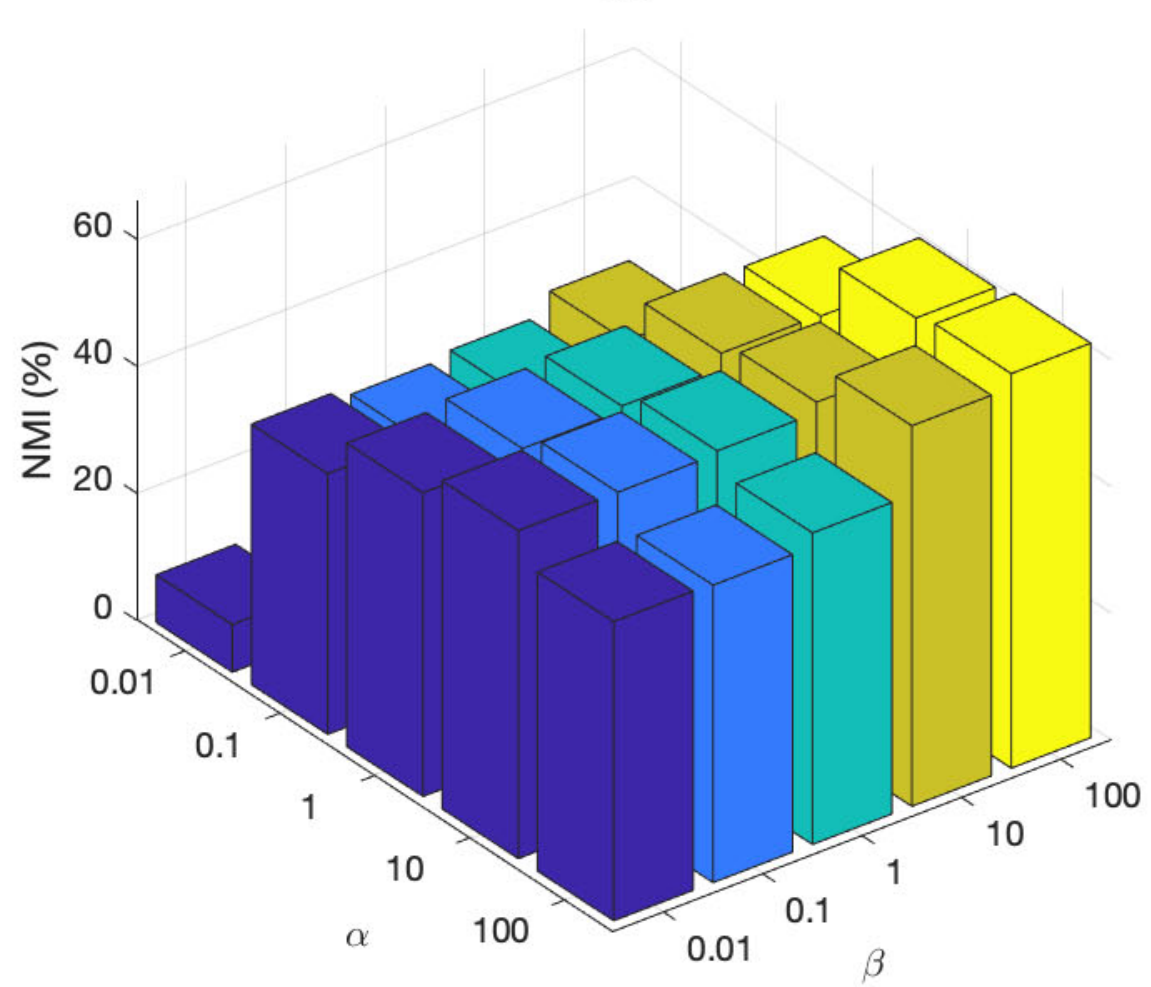}}}
		\caption{Performance of the proposed JSMC method with varying parameters $\alpha$ and $\beta$ on the benchmark datasets.}
		\label{fig:lambda}
	\end{center}
\end{figure*}

\begin{figure*}[!t]
	\begin{center}
		{\subfigure[\emph{3Sources}]
			{\includegraphics[width=0.23\columnwidth]{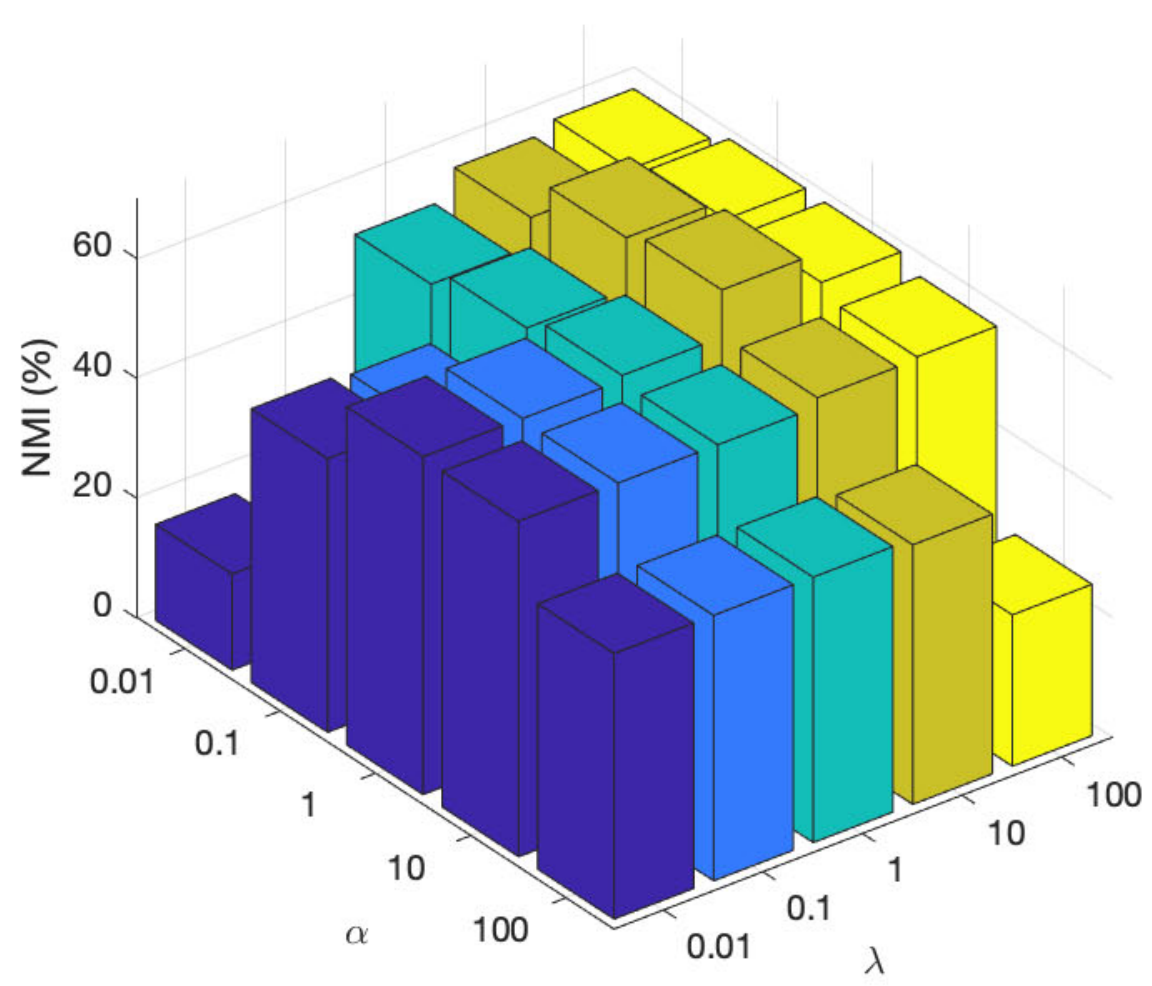}}}
		{\subfigure[\emph{Notting-Hill}]
			{\includegraphics[width=0.23\columnwidth]{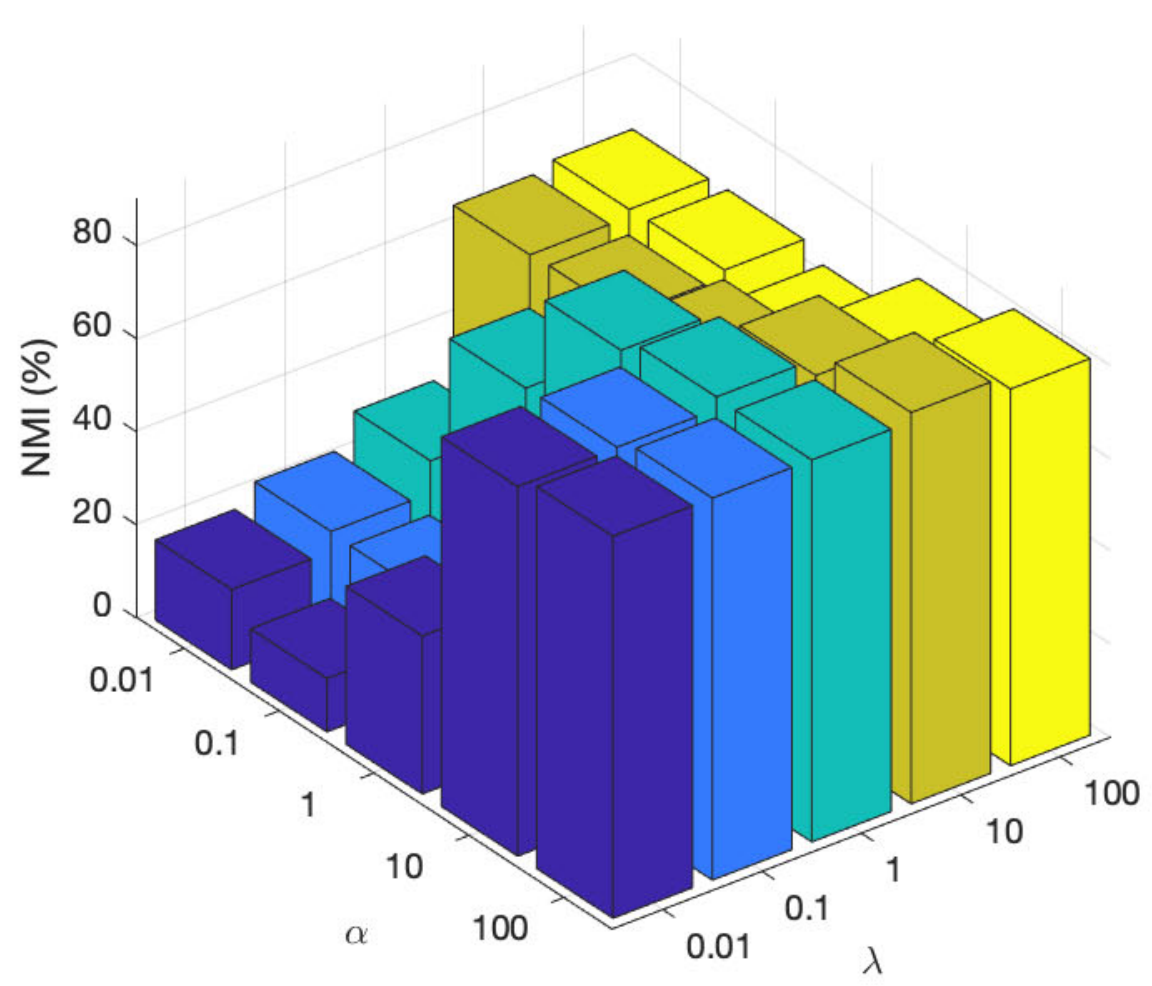}}}
		{\subfigure[\emph{ORL}]
			{\includegraphics[width=0.23\columnwidth]{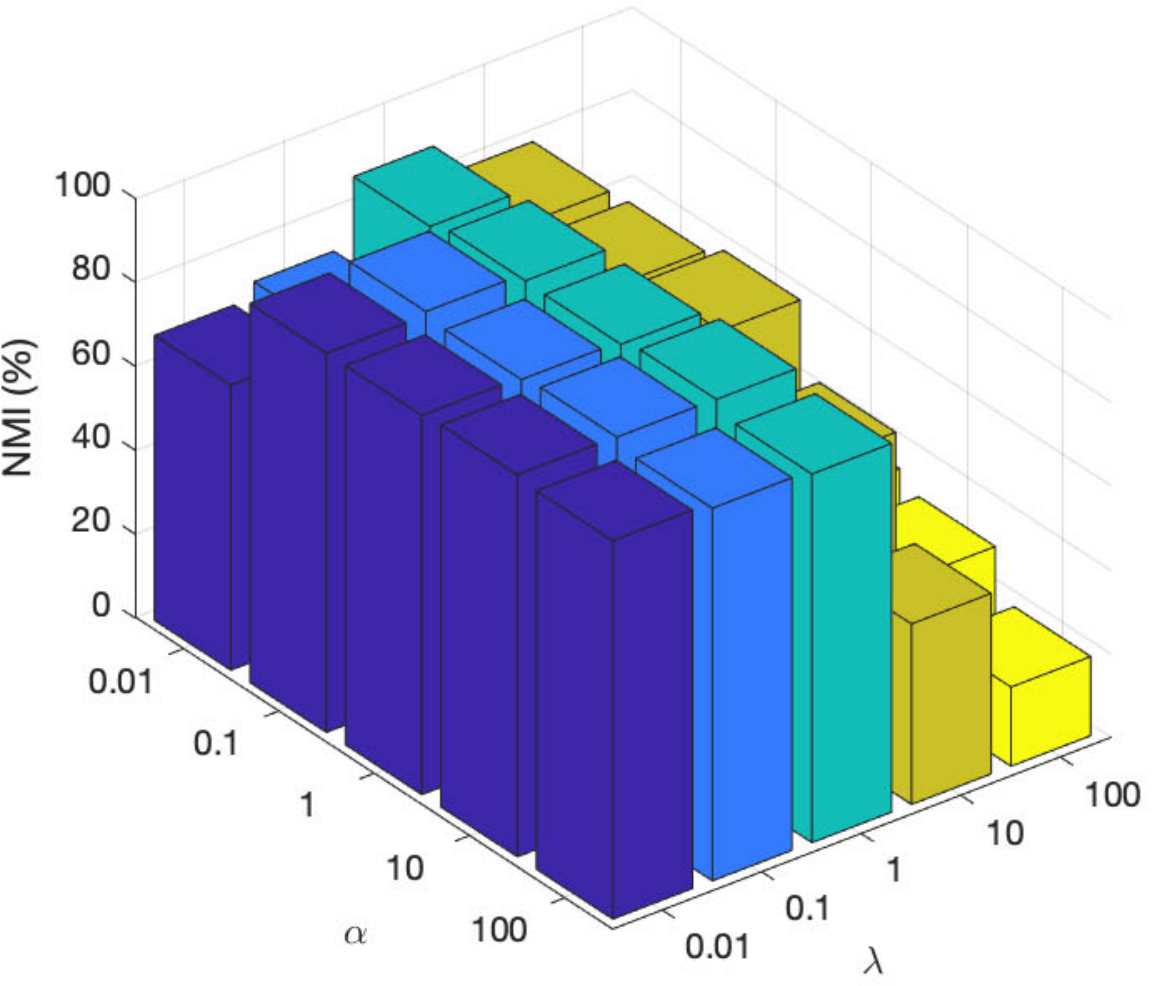}}}
		{\subfigure[\emph{WebKB-Texas}]
			{\includegraphics[width=0.23\columnwidth]{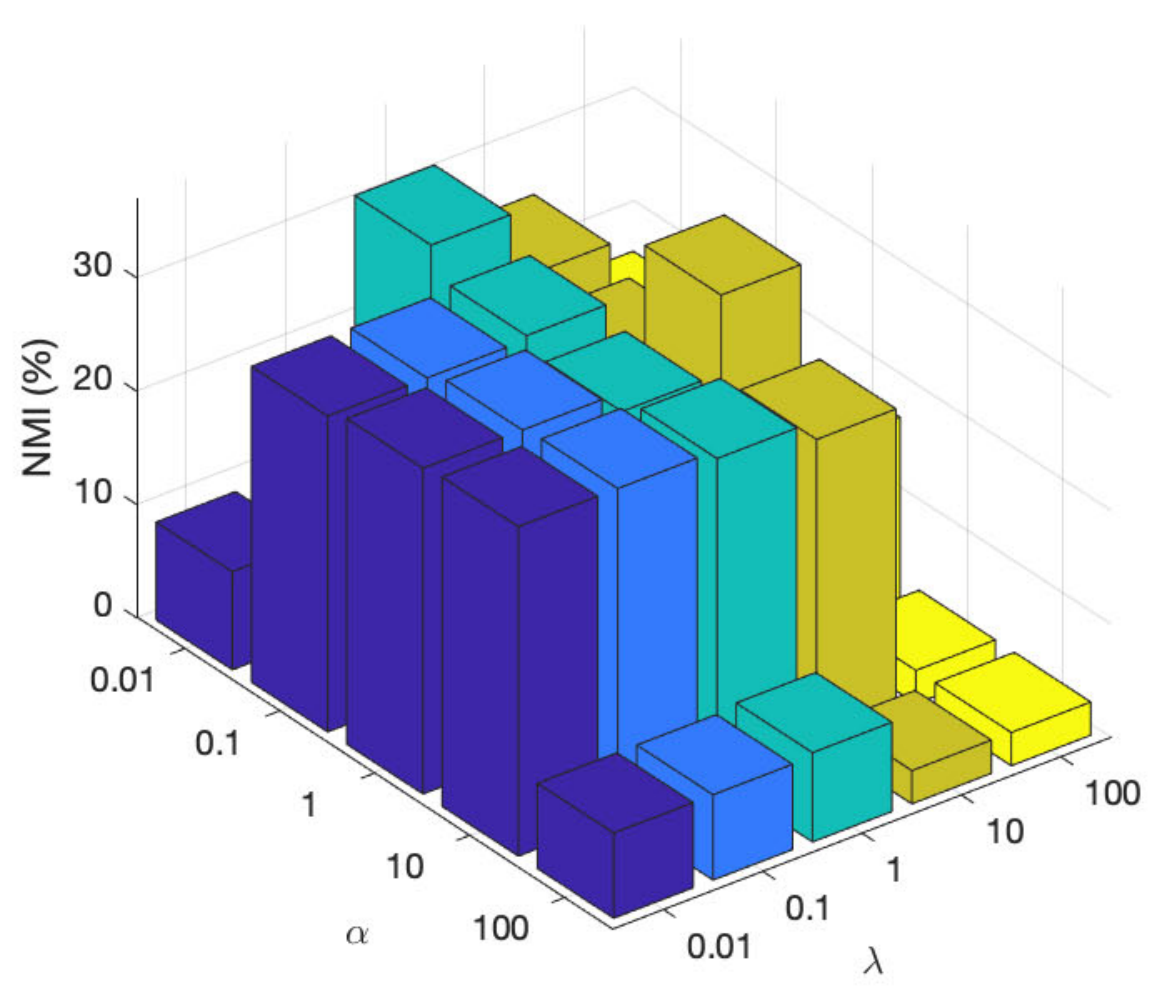}}}
		{\subfigure[\emph{Yale}]
			{\includegraphics[width=0.23\columnwidth]{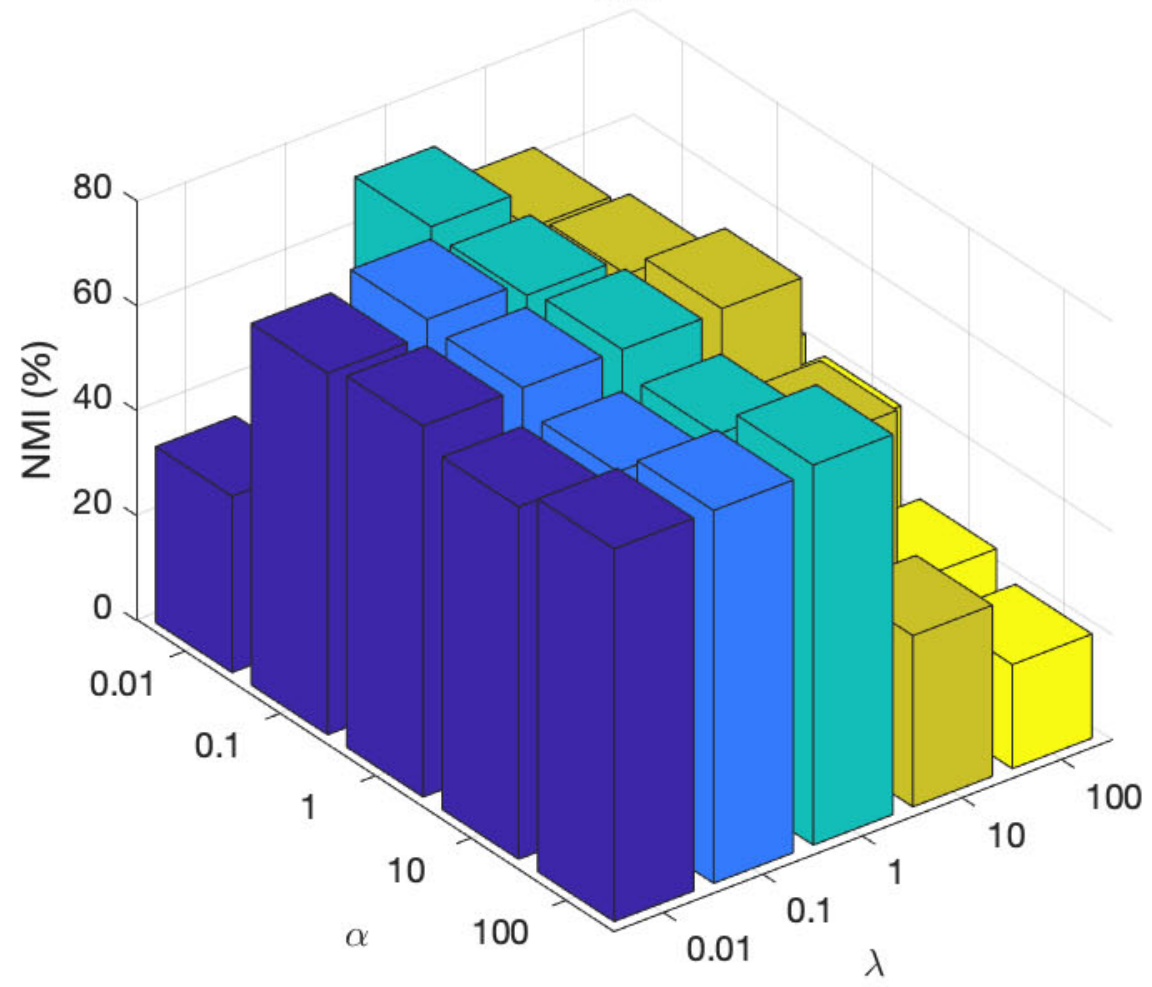}}}
		{\subfigure[\emph{Reuters}]
			{\includegraphics[width=0.23\columnwidth]{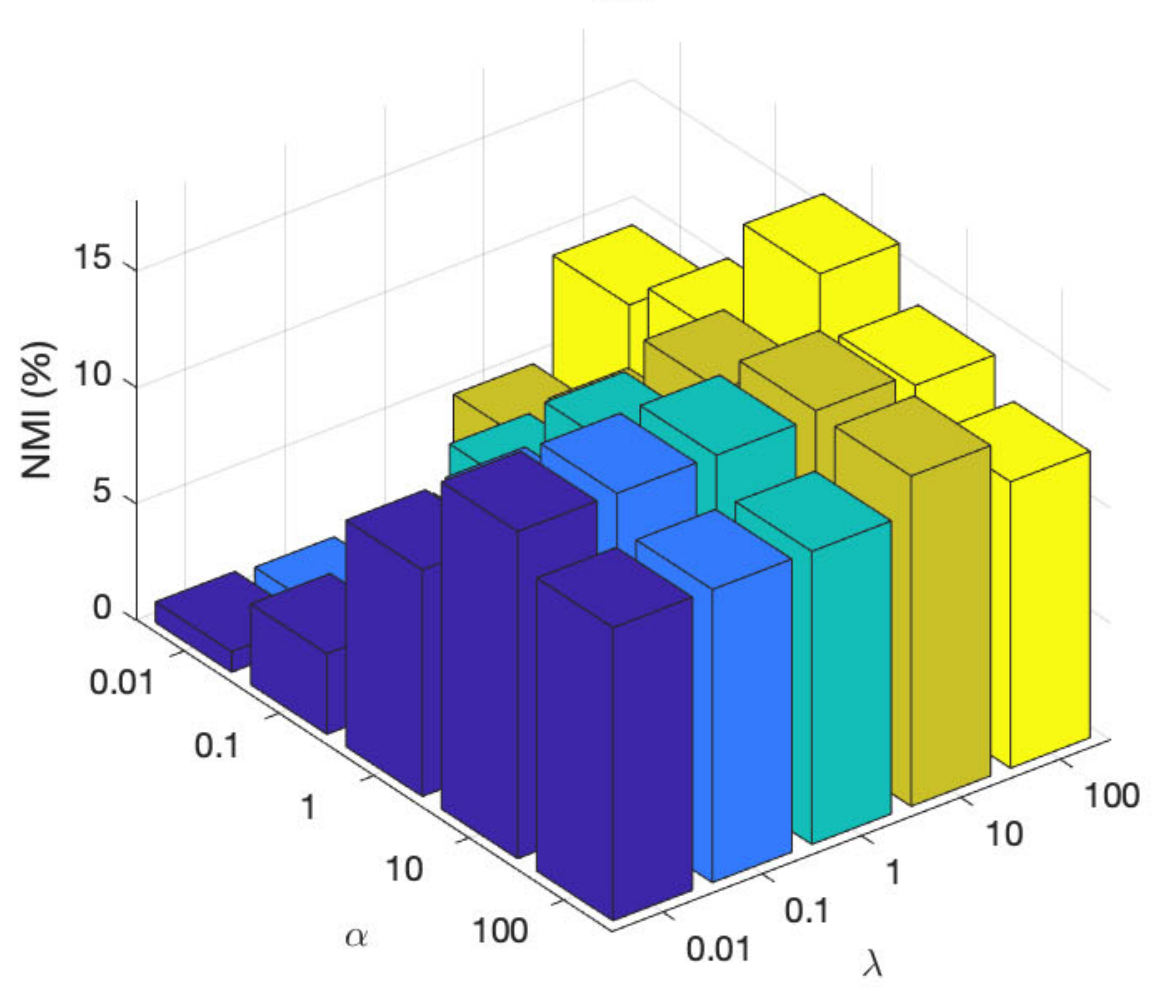}}}
		{\subfigure[\emph{COIL-20}]
			{\includegraphics[width=0.23\columnwidth]{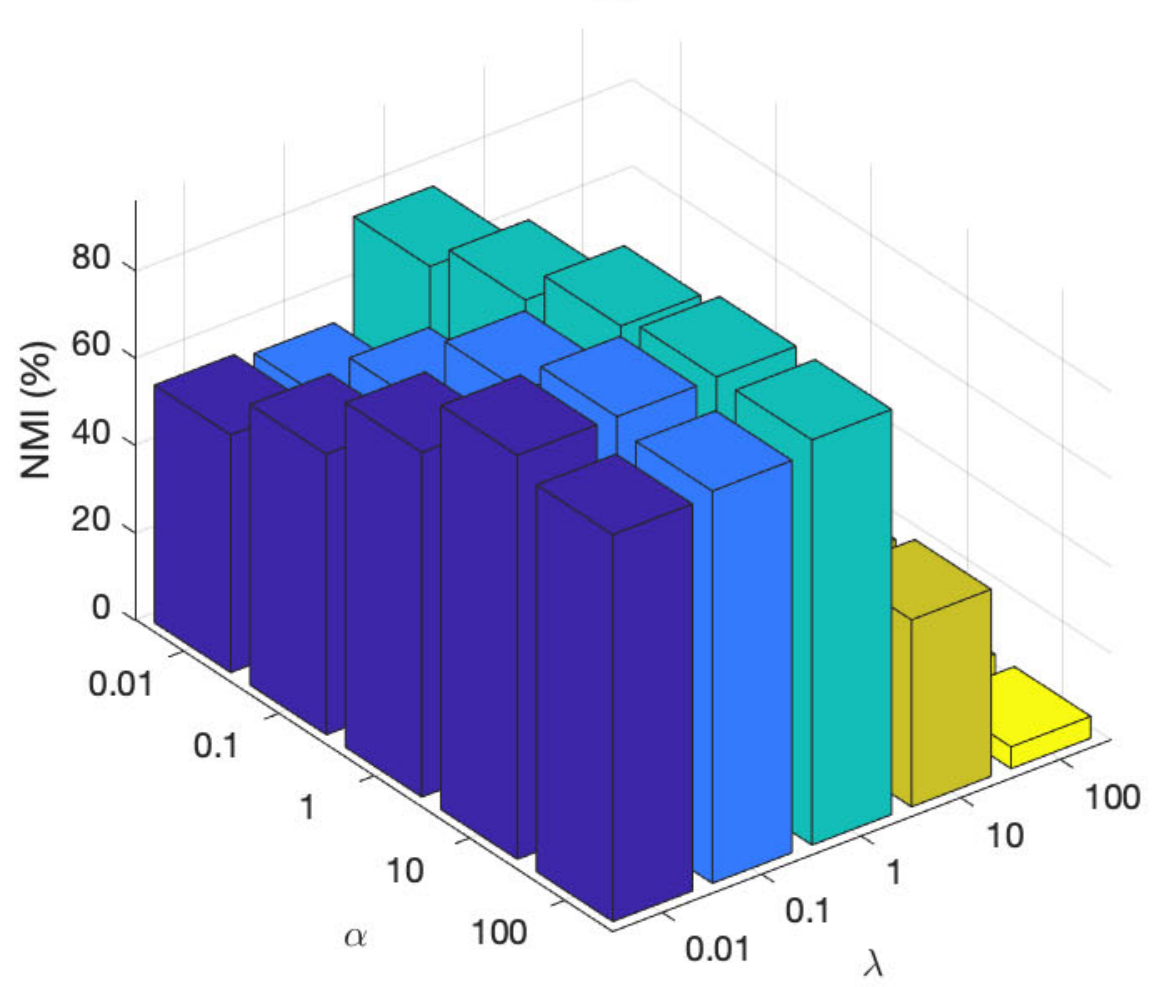}}}
		{\subfigure[\emph{Caltech-7}]
			{\includegraphics[width=0.23\columnwidth]{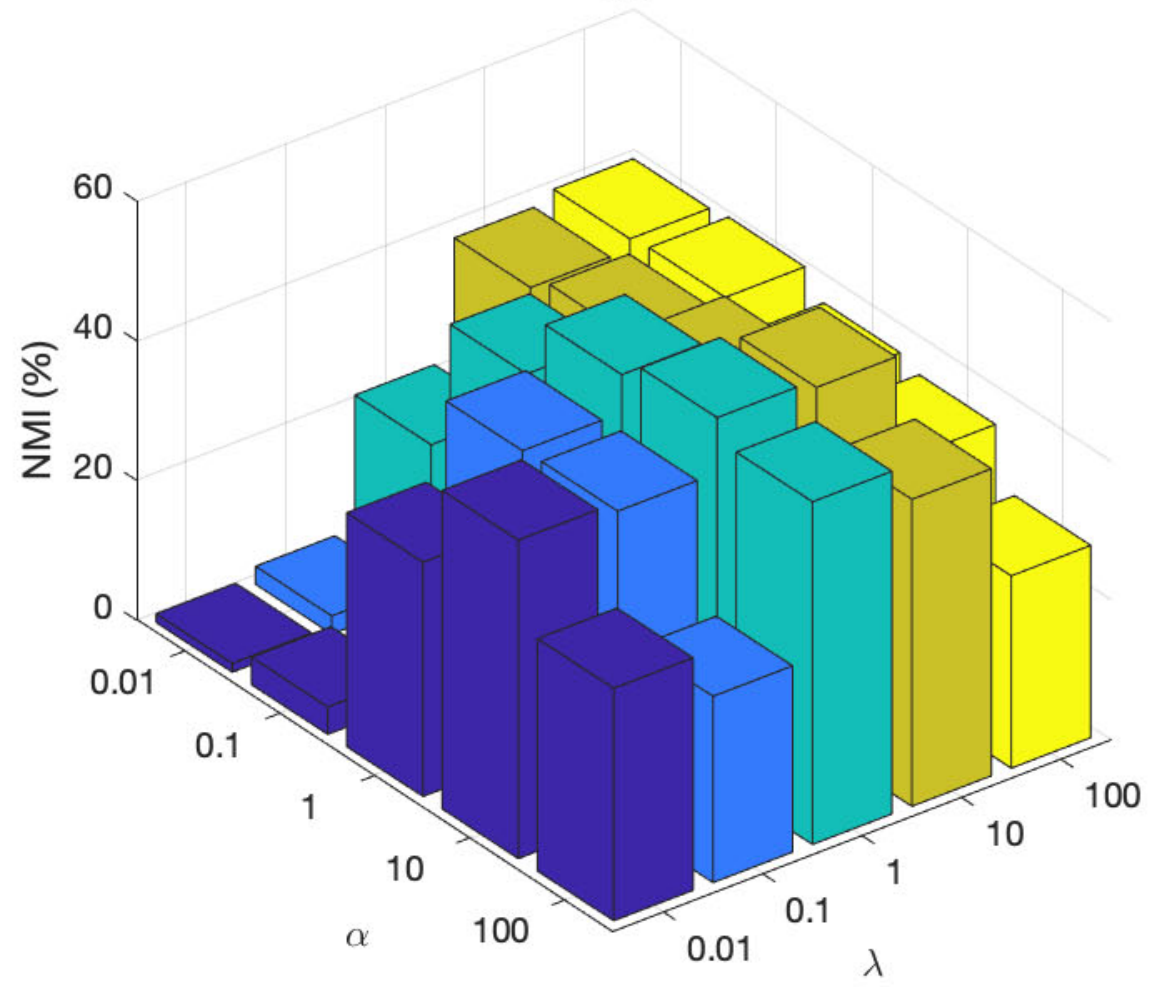}}}
		\caption{Performance of the proposed JSMC method with varying parameters $\alpha$ and $\lambda$ on the benchmark datasets.}
		\label{fig:beta}
	\end{center}
\end{figure*}

In this section, we evaluate the performance (w.r.t. NMI(\%)) of the proposed JSMC method with varying parameter settings on the benchmark datasets. Specifically, we first evaluate the performance of JSMC with varying parameters $\alpha$ and $\beta$, and illustrate the performance results in Fig.~\ref{fig:lambda}. Then we evaluate the performance of JSMC with varying parameters  $\alpha$ and $\lambda$, and illustrate the performance results in Fig.~\ref{fig:beta}. As shown in Figs.~\ref{fig:lambda}, and~\ref{fig:beta}, the proposed JSMC method has shown relative consistent performance with varying parameter settings. Especially, setting the three parameters, i.e., $\alpha$, $\beta$, and $\lambda$, to moderate values can generally lead to favorable clustering results on the benchmark datasets.

\begin{figure*}[!t]
	\begin{center}
		{\subfigure[\emph{3Sources}]
			{\includegraphics[width=0.243\columnwidth]{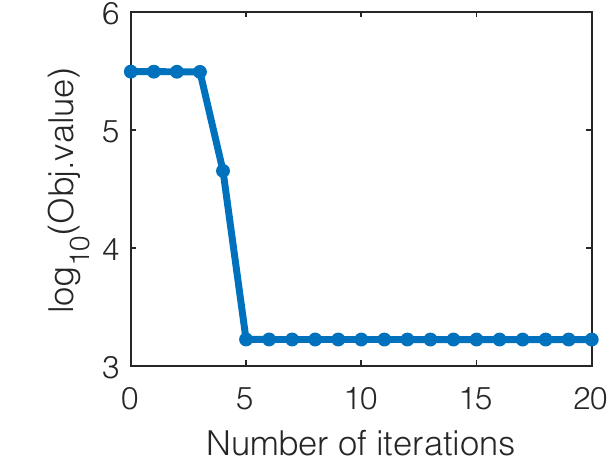}}}
		{\subfigure[\emph{Notting-Hill}]
			{\includegraphics[width=0.243\columnwidth]{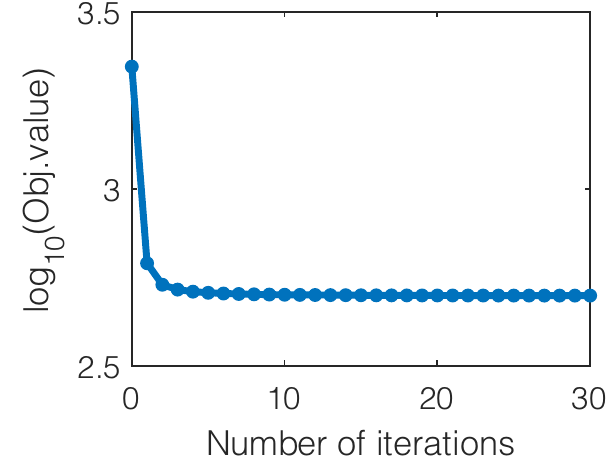}}}
		{\subfigure[\emph{ORL}]
			{\includegraphics[width=0.243\columnwidth]{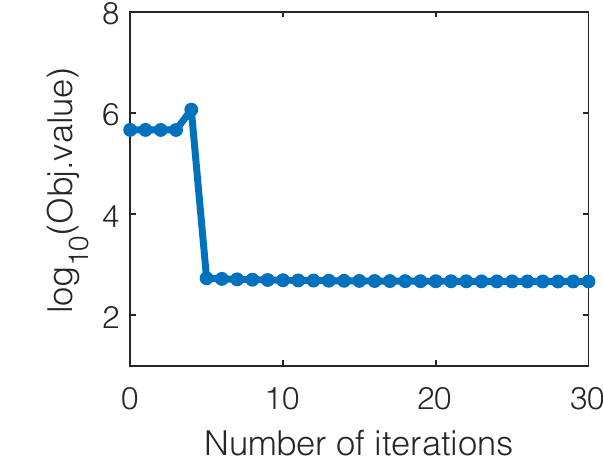}}}
		{\subfigure[\emph{WebKB-Texas}]
			{\includegraphics[width=0.243\columnwidth]{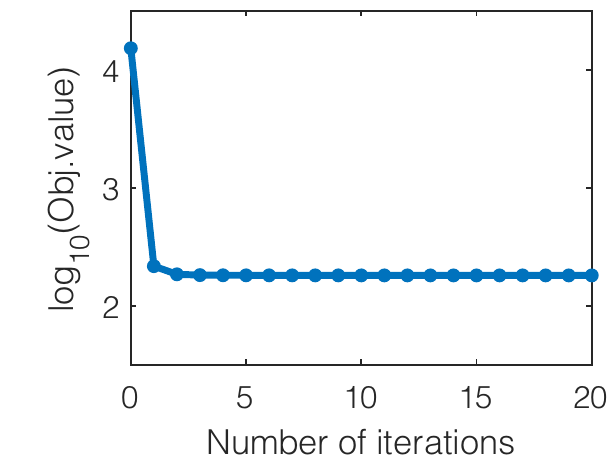}}}
		{\subfigure[\emph{Yale}]
			{\includegraphics[width=0.243\columnwidth]{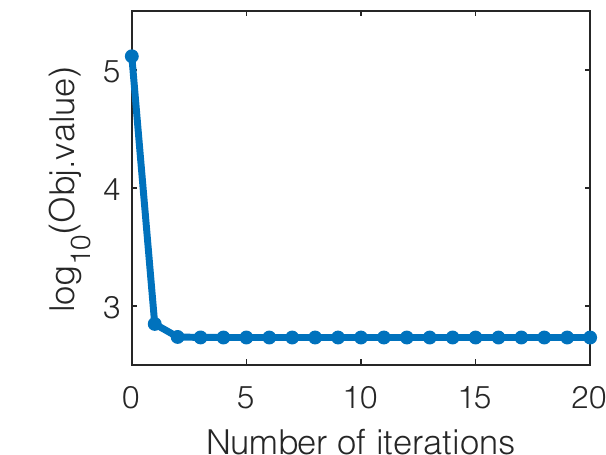}}}
		{\subfigure[\emph{Reuters}]
			{\includegraphics[width=0.243\columnwidth]{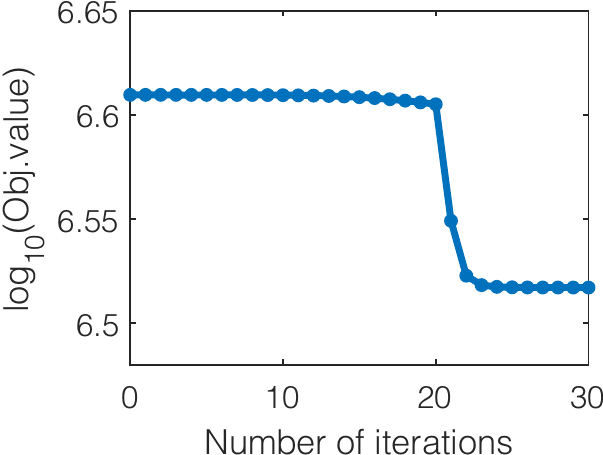}}}
		{\subfigure[\emph{COIL-20}]
			{\includegraphics[width=0.243\columnwidth]{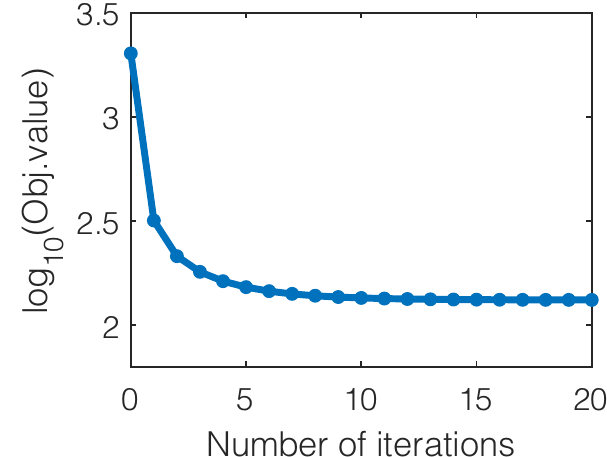}}}
		{\subfigure[\emph{Caltech-7}]
			{\includegraphics[width=0.243\columnwidth]{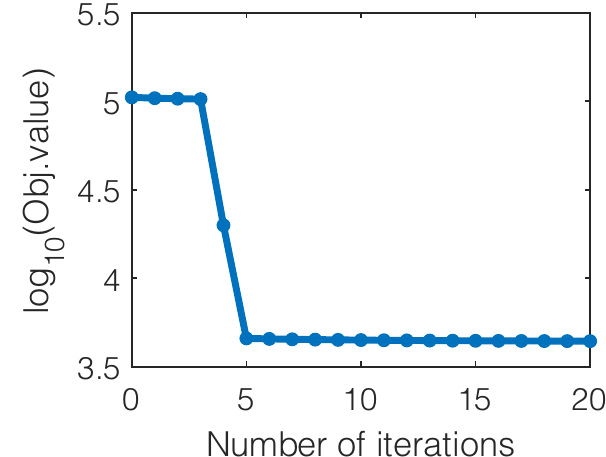}}}
		\caption{Convergence of the objective function~(\ref{eq:ObjFunction})  with increasing iterations on the eight benchmark datasets.}
		\label{fig:con}
	\end{center}
\end{figure*}

\subsection{Convergence Analysis}
\label{sec:convergence}
In this section, we analyze the convergence of the proposed JSMC method on the benchmark datasets. In JSMC, the objective function~(\ref{eq:ObjFunction}) is formulated with the cross-view consistency and inconsistency, the view-consensus grouping effect, and the low-rank representation simutaneously considered, which is then optimized by an alternating optimization algorithm.
As shown in Fig.~\ref{fig:con}, the convergence of the objective function~(\ref{eq:ObjFunction}) is generally reached within thirty iterations, which indicates a fast convergence speed of the proposed method on a variety of datasets.

\subsection{Ablation Study}
\label{sec:ablation}
In this section, the ablation study is conducted to test the influences of different terms (or components) in our objective function~(\ref{eq:ObjFunction}). Specifically, the influences of three terms, i.e., the inconsistency term, the smoothness term (via the grouping effect), and the low-rank regularization term, are evluated, with the evaluation results w.r.t. NMI, ARI, ACC, and PUR reported in Tables~\ref{tab:Ablation_nmi},~\ref{tab:Ablation_ari},~\ref{tab:Ablation_acc},~\ref{tab:Ablation_purity}, respectively. In each of these four tables, the first column corresponds to the  results of JSMC with all the three terms (or components). The second, third, and fourth columns correspond to the results with one component removed. And the fifth, sixth, and seventh columns correspond to the results with two components removed.

As shown in Tables~\ref{tab:Ablation_nmi}, the average NMI(\%) score across eight datasets is 69.31. When removing each of the three components, the average NMI(\%) scores decrease to 61.75, 59.78, and 65.24, respectively, which demonstrate the advantageous contributions of the three components. Moreover, when removing two of the three components and only preserving one of them, the average NMI(\%) scores further decrease to 36.93, 60.66, and 7.83, respectively. Similar influences of the three components over the clustering performance w.r.t. ARI(\%), ACC(\%), PUR(\%) can also be observed in Tables~\ref{tab:Ablation_ari},~\ref{tab:Ablation_acc},~\ref{tab:Ablation_purity}, repectively, which have shown the influence of each component and the joint benefits of formulating these three components into a unified model.

\section{Conclusion and Future Work}
\label{sec:conclusion}

\begin{table*}[!t]
	\caption{Ablation analysis (w.r.t. NMI(\%)) on the benchmark datasets. The best score in each row is highlighted in bold.}
	\label{tab:Ablation_nmi}\vskip 0.02 in
	\renewcommand\arraystretch{1.25}
	\scalebox{0.65}{
		\centering
		\begin{tabular}{|m{2.4cm}<{\centering}|m{2cm}<{\centering}|m{1.8cm}<{\centering}m{1.8cm}<{\centering}m{1.8cm}<{\centering}|m{1.8cm}<{\centering}m{1.8cm}<{\centering}m{1.8cm}<{\centering}|}
			\hline
			Method        &JSMC     &\multicolumn{3}{c|}{Removing One Component}    &\multicolumn{3}{c|}{Removing Two Components}    \\
			\hline
			Inconsistency &\checkmark    &  &\checkmark   &\checkmark   & & &\checkmark  \\
			Smoothness        &\checkmark    &\checkmark    & &\checkmark   & &\checkmark   & \\
			Low-rank      &\checkmark    &\checkmark    &\checkmark   & &\checkmark   & & \\
			\hline
			\emph{3Sources}		&\textbf{69.52}$_{\pm0.00}$	&57.88$_{\pm0.00}$	&\textbf{69.52}$_{\pm0.00}$	&58.77$_{\pm0.00}$	&54.06$_{\pm0.00}$	&57.88$_{\pm0.00}$	&25.97$_{\pm0.00}$\\
			\emph{Notting-Hill}		&\textbf{96.92}$_{\pm0.00}$	&81.86$_{\pm0.00}$	&66.45$_{\pm0.00}$	&\textbf{96.92}$_{\pm0.00}$	&16.89$_{\pm0.00}$	&81.86$_{\pm0.00}$	&0.66$_{\pm0.00}$\\
			\emph{ORL}		&\textbf{91.46}$_{\pm0.00}$	&88.65$_{\pm0.00}$	&88.64$_{\pm0.00}$	&91.41$_{\pm0.00}$	&77.64$_{\pm0.00}$	&89.19$_{\pm0.00}$	&25.55$_{\pm0.00}$\\
			\emph{WebKB-Texas}		&\textbf{38.04}$_{\pm0.00}$	&30.44$_{\pm0.00}$	&33.72$_{\pm0.00}$	&30.63$_{\pm0.00}$	&17.55$_{\pm0.00}$	&29.87$_{\pm0.00}$	&3.51$_{\pm0.00}$\\
			\emph{Yale}		&\textbf{76.38}$_{\pm0.00}$	&76.01$_{\pm0.00}$	&70.08$_{\pm0.00}$	&72.85$_{\pm0.00}$	&59.03$_{\pm0.00}$	&69.41$_{\pm0.00}$	&0.00$_{\pm0.00}$\\
			\emph{Reuters}		&\textbf{25.96}$_{\pm0.00}$	&15.14$_{\pm0.00}$	&25.92$_{\pm0.00}$	&15.05$_{\pm0.00}$	&3.05$_{\pm0.00}$	&15.14$_{\pm0.00}$	&0.00$_{\pm0.00}$\\
			\emph{COIL-20}		&92.98$_{\pm0.00}$	&89.56$_{\pm0.00}$	&82.06$_{\pm0.00}$	&\textbf{93.19}$_{\pm0.00}$	&65.62$_{\pm0.00}$	&90.01$_{\pm0.00}$	&6.00$_{\pm0.00}$\\
			\emph{Caltech-7}		&\textbf{63.20}$_{\pm0.00}$	&54.42$_{\pm0.00}$	&41.83$_{\pm0.00}$	&63.09$_{\pm0.00}$	&1.61$_{\pm0.00}$	&51.91$_{\pm0.00}$	&0.92$_{\pm0.00}$\\
			\hline
			Avg. score		&\textbf{69.31}	&61.75	&59.78	&65.24	&36.93	&60.66	&7.83\\
			\hline
		\end{tabular}
	}
\end{table*}

\begin{table*}[!t]
	\caption{Ablation analysis (w.r.t. ARI(\%)) on the benchmark datasets. The best score in each row is highlighted in bold.}
	\label{tab:Ablation_ari}\vskip 0.02 in
	\renewcommand\arraystretch{1.25}
	\scalebox{0.65}{
		\centering
		\begin{tabular}{|m{2.4cm}<{\centering}|m{2cm}<{\centering}|m{1.8cm}<{\centering}m{1.8cm}<{\centering}m{1.8cm}<{\centering}|m{1.8cm}<{\centering}m{1.8cm}<{\centering}m{1.8cm}<{\centering}|}
			\hline
			Method        &JSMC     &\multicolumn{3}{c|}{Removing One Component}    &\multicolumn{3}{c|}{Removing Two Components}    \\
			\hline
			Inconsistency &\checkmark    &  &\checkmark   &\checkmark   & & &\checkmark  \\
			Smoothness        &\checkmark    &\checkmark    & &\checkmark   & &\checkmark   & \\
			Low-rank      &\checkmark    &\checkmark    &\checkmark   & &\checkmark   & & \\
			\hline
			\emph{3Sources}		&\textbf{65.17}$_{\pm0.00}$	&42.30$_{\pm0.00}$	&\textbf{65.17}$_{\pm0.00}$	&43.03$_{\pm0.00}$	&43.27$_{\pm0.00}$	&42.30$_{\pm0.00}$	&19.86$_{\pm0.00}$\\
			\emph{Notting-Hill}		&\textbf{97.69}$_{\pm0.00}$	&80.25$_{\pm0.00}$	&66.72$_{\pm0.00}$	&\textbf{97.69}$_{\pm0.00}$	&16.21$_{\pm0.00}$	&80.25$_{\pm0.00}$	&-0.29$_{\pm0.00}$\\
			\emph{ORL}		&\textbf{77.79}$_{\pm0.00}$	&70.81$_{\pm0.00}$	&70.15$_{\pm0.00}$	&76.82$_{\pm0.00}$	&48.24$_{\pm0.00}$	&71.81$_{\pm0.00}$	&1.23$_{\pm0.00}$\\
			\emph{WebKB-Texas}		&\textbf{32.26}$_{\pm0.00}$	&19.94$_{\pm0.00}$	&26.79$_{\pm0.00}$	&21.38$_{\pm0.00}$	&15.61$_{\pm0.00}$	&21.19$_{\pm0.00}$	&3.64$_{\pm0.00}$\\
			\emph{Yale}		&\textbf{57.58}$_{\pm0.00}$	&57.13$_{\pm0.00}$	&50.85$_{\pm0.00}$	&54.73$_{\pm0.00}$	&33.37$_{\pm0.00}$	&49.06$_{\pm0.00}$	&0.00$_{\pm0.00}$\\
			\emph{Reuters}		&\textbf{20.48}$_{\pm0.00}$	&5.34$_{\pm0.00}$	&20.42$_{\pm0.00}$	&5.10$_{\pm0.00}$	&2.29$_{\pm0.00}$	&5.34$_{\pm0.00}$	&0.00$_{\pm0.00}$\\
			\emph{COIL-20}		&81.28$_{\pm0.00}$	&79.61$_{\pm0.00}$	&68.79$_{\pm0.00}$	&\textbf{81.64}$_{\pm0.00}$	&44.55$_{\pm0.00}$	&79.03$_{\pm0.00}$	&1.73$_{\pm0.00}$\\
			\emph{Caltech-7}		&\textbf{60.10}$_{\pm0.00}$	&48.42$_{\pm0.00}$	&35.56$_{\pm0.00}$	&60.00$_{\pm0.00}$	&0.97$_{\pm0.00}$	&47.35$_{\pm0.00}$	&0.96$_{\pm0.00}$\\
			\hline
			Avg. score		&\textbf{61.54}	&50.48	&50.56	&55.05	&25.56	&49.54	&3.39\\
			\hline
		\end{tabular}
	}
\end{table*}

In this paper, we present a novel multi-view subspace clustering approach termed JSMC. In the proposed approach, to simultaneously capture  the cross-view commonness and inconsistencies, we decompose the subspace representation on each view into two matrices, i.e., the view-commonness matrix and the view-inconsistency matrix. The view-commonness matrix is further constrained by the view-consensus grouping effect, which exploits the multi-view local structures to promote the learning of the common representation. Moreover, the low-rank representation is incorporated via the nuclear norm to strengthen the cluster structure and improve the clustering robustness.  Thereby, the cross-view commonness and inconsistencies, the view-consensus grouping effect, and the low-rank representation are jointly leveraged in a unified objective function, which is optimized via an alternating minimization algorithm. Experimental results on a variety of multi-view datasets have demonstrated the superior performance of the proposed JSMC approach.

In this paper, we mainly focus on the effectiveness and robustness of multi-view clustering. In the future work, we plan to extend our JSMC approach from the general graph formulation to the bipartite graph formulation \cite{sun2021scalable} so as to make it feasible for large-scale applications. Furthermore, different from building a single clustering result for a dataset, it may also be a promising direction to extend the proposed approach to an ensemble clustering framework \cite{Huang2019,Huang2021a} or a multiple clustering framework \cite{yao19_ijcai} in the future research.

\section*{Acknowledgments}
This work was supported by the NSFC (61976097, 62206096 \& 62276277), the Natural Science Foundation of Guangdong Province (2021A1515012203), and the Science and Technology Program of Guangzhou, China (202201010314).

\begin{table*}[!t]
	\caption{Ablation analysis (w.r.t. ACC(\%)) on the benchmark datasets. The best score in each row is highlighted in bold.}
	\label{tab:Ablation_acc}\vskip 0.02 in
	\renewcommand\arraystretch{1.25}
	\scalebox{0.65}{
		\centering
		\begin{tabular}{|m{2.4cm}<{\centering}|m{2cm}<{\centering}|m{1.8cm}<{\centering}m{1.8cm}<{\centering}m{1.8cm}<{\centering}|m{1.8cm}<{\centering}m{1.8cm}<{\centering}m{1.8cm}<{\centering}|}
			\hline
			Method        &JSMC     &\multicolumn{3}{c|}{Removing One Component}    &\multicolumn{3}{c|}{Removing Two Components}    \\
			\hline
			Inconsistency &\checkmark    &  &\checkmark   &\checkmark   & & &\checkmark  \\
			Smoothness        &\checkmark    &\checkmark    & &\checkmark   & &\checkmark   & \\
			Low-rank      &\checkmark    &\checkmark    &\checkmark   & &\checkmark   & & \\
			\hline
			\emph{3Sources}		&\textbf{77.51}$_{\pm0.00}$	&62.13$_{\pm0.00}$	&\textbf{77.51}$_{\pm0.00}$	&62.13$_{\pm0.00}$	&52.66$_{\pm0.00}$	&62.13$_{\pm0.00}$	&43.20$_{\pm0.00}$\\
			\emph{Notting-Hill}		&\textbf{98.91}$_{\pm0.00}$	&84.91$_{\pm0.00}$	&79.09$_{\pm0.00}$	&\textbf{98.91}$_{\pm0.00}$	&47.64$_{\pm0.00}$	&84.91$_{\pm0.00}$	&24.36$_{\pm0.00}$\\
			\emph{ORL}		&\textbf{82.00}$_{\pm0.00}$	&79.75$_{\pm0.00}$	&75.50$_{\pm0.00}$	&81.50$_{\pm0.00}$	&61.50$_{\pm0.00}$	&78.75$_{\pm0.00}$	&14.25$_{\pm0.00}$\\
			\emph{WebKB-Texas}		&\textbf{59.89}$_{\pm0.00}$	&50.27$_{\pm0.00}$	&55.08$_{\pm0.00}$	&51.34$_{\pm0.00}$	&46.52$_{\pm0.00}$	&51.87$_{\pm0.00}$	&43.85$_{\pm0.00}$\\
			\emph{Yale}		&\textbf{73.33}$_{\pm0.00}$	&70.30$_{\pm0.00}$	&68.48$_{\pm0.00}$	&69.70$_{\pm0.00}$	&50.91$_{\pm0.00}$	&67.27$_{\pm0.00}$	&0.00$_{\pm0.00}$\\
			\emph{Reuters}		&\textbf{46.75}$_{\pm0.00}$	&30.67$_{\pm0.00}$	&46.67$_{\pm0.00}$	&30.42$_{\pm0.00}$	&24.58$_{\pm0.00}$	&30.67$_{\pm0.00}$	&0.00$_{\pm0.00}$\\
			\emph{COIL-20}		&83.96$_{\pm0.00}$	&81.88$_{\pm0.00}$	&73.96$_{\pm0.00}$	&\textbf{84.24}$_{\pm0.00}$	&55.76$_{\pm0.00}$	&80.49$_{\pm0.00}$	&9.86$_{\pm0.00}$\\
			\emph{Caltech-7}		&70.90$_{\pm0.00}$	&61.47$_{\pm0.00}$	&49.66$_{\pm0.00}$	&\textbf{71.51}$_{\pm0.00}$	&20.01$_{\pm0.00}$	&63.43$_{\pm0.00}$	&50.07$_{\pm0.00}$\\
			\hline
			Avg. score		&\textbf{74.16}	&65.17	&65.74	&68.72	&44.95	&64.94	&23.20\\
			\hline
		\end{tabular}
	}
\end{table*}

\begin{table*}[!t]
	\caption{Ablation analysis (w.r.t. PUR(\%)) on the benchmark datasets. The best score in each row is highlighted in bold.}
	\label{tab:Ablation_purity}\vskip 0.02 in
	\renewcommand\arraystretch{1.25}
	\scalebox{0.65}{
		\centering
		\begin{tabular}{|m{2.4cm}<{\centering}|m{2cm}<{\centering}|m{1.8cm}<{\centering}m{1.8cm}<{\centering}m{1.8cm}<{\centering}|m{1.8cm}<{\centering}m{1.8cm}<{\centering}m{1.8cm}<{\centering}|}
			\hline
			Method        &JSMC     &\multicolumn{3}{c|}{Removing One Component}    &\multicolumn{3}{c|}{Removing Two Components}    \\
			\hline
			Inconsistency &\checkmark    &  &\checkmark   &\checkmark   & & &\checkmark  \\
			Smoothness        &\checkmark    &\checkmark    & &\checkmark   & &\checkmark   & \\
			Low-rank      &\checkmark    &\checkmark    &\checkmark   & &\checkmark   & & \\
			\hline
			\emph{3Sources}		&\textbf{82.25}$_{\pm0.00}$	&73.96$_{\pm0.00}$	&\textbf{82.25}$_{\pm0.00}$	&74.56$_{\pm0.00}$	&69.82$_{\pm0.00}$	&73.96$_{\pm0.00}$	&58.58$_{\pm0.00}$\\
			\emph{Notting-Hill}		&\textbf{98.91}$_{\pm0.00}$	&86.36$_{\pm0.00}$	&80.36$_{\pm0.00}$	&\textbf{98.91}$_{\pm0.00}$	&49.09$_{\pm0.00}$	&86.36$_{\pm0.00}$	&31.27$_{\pm0.00}$\\
			\emph{ORL}		&84.50$_{\pm0.00}$	&82.00$_{\pm0.00}$	&79.50$_{\pm0.00}$	&\textbf{85.25}$_{\pm0.00}$	&65.50$_{\pm0.00}$	&81.75$_{\pm0.00}$	&14.50$_{\pm0.00}$\\
			\emph{WebKB-Texas}		&70.05$_{\pm0.00}$	&67.38$_{\pm0.00}$	&\textbf{70.59}$_{\pm0.00}$	&68.45$_{\pm0.00}$	&63.10$_{\pm0.00}$	&67.91$_{\pm0.00}$	&55.08$_{\pm0.00}$\\
			\emph{Yale}		&\textbf{73.94}$_{\pm0.00}$	&70.91$_{\pm0.00}$	&68.48$_{\pm0.00}$	&70.30$_{\pm0.00}$	&53.33$_{\pm0.00}$	&68.48$_{\pm0.00}$	&0.00$_{\pm0.00}$\\
			\emph{Reuters}		&\textbf{48.58}$_{\pm0.00}$	&33.25$_{\pm0.00}$	&48.50$_{\pm0.00}$	&33.00$_{\pm0.00}$	&25.08$_{\pm0.00}$	&33.25$_{\pm0.00}$	&0.00$_{\pm0.00}$\\
			\emph{COIL-20}		&88.68$_{\pm0.00}$	&84.24$_{\pm0.00}$	&75.63$_{\pm0.00}$	&\textbf{88.89}$_{\pm0.00}$	&58.06$_{\pm0.00}$	&84.31$_{\pm0.00}$	&9.93$_{\pm0.00}$\\
			\emph{Caltech-7}		&\textbf{92.06}$_{\pm0.00}$	&89.15$_{\pm0.00}$	&84.19$_{\pm0.00}$	&90.91$_{\pm0.00}$	&54.14$_{\pm0.00}$	&87.25$_{\pm0.00}$	&54.55$_{\pm0.00}$\\
			\hline
			Avg. score		&\textbf{79.87}	&73.41	&73.69	&76.28	&54.76	&72.91	&27.99\\
			\hline
		\end{tabular}
	}
\end{table*}

\bibliographystyle{elsarticle-num-names}
\bibliography{refs_2021}

\newpage

\end{document}